\documentclass[11pt]{article}

\usepackage[final]{acl}

\usepackage{times}
\usepackage{latexsym}

\usepackage[T1]{fontenc}

\usepackage[utf8]{inputenc}

\usepackage{microtype}

\usepackage{inconsolata}
\usepackage{graphicx}
\usepackage{booktabs}
\usepackage{multirow}

\usepackage{afterpage}
\usepackage{float}
\usepackage{amsmath}
\usepackage{amsfonts}
\usepackage{changepage}
\usepackage[table]{xcolor}

\newcommand{\vcd}{~\cite{leng2024mitigating_vcd}}
\newcommand{\md}{~\cite{favero2024mitigating_m3id}}
\newcommand{\ritual}{~\cite{woo2024ritual}}
\newcommand{\degf}{~\cite{zhang2025degf}}
\newcommand{\llava}{~\cite{liu2023improvedllava}}
\newcommand{\insblip}{~\cite{dai2023instructblip}}

\title{MRFD: Multi-Region Fusion Decoding with Self-Consistency for Mitigating Hallucinations in LVLMs}

\author{
\textbf{Haonan Ge}$^{\lambda}$ \quad
\textbf{Yiwei Wang}$^{\lambda}$\thanks{\, Corresponding author.} \quad
\textbf{Ming-Hsuan Yang}$^{\lambda}$ \quad
\textbf{Yujun Cai}$^{\dagger}$ \\
$^{\lambda}$ Department of Computer Science and Engineering, University of California at Merced \\
$^{\dagger}$ The University of Queensland \\
\texttt{gehaonan82@gmail.com}\quad
\texttt{wangyw.evan@gmail.com}\\
\href{https://mrfd1.github.io/}{\texttt{https://mrfd1.github.io}}
}

\begin{document}
\maketitle

\begin{abstract}
Large Vision-Language Models (LVLMs) have shown strong performance across multimodal tasks. However, they often produce hallucinations—text that is inconsistent with visual input, due to the limited ability to verify information in different regions of the image. To address this, we propose \textbf{Multi-Region Fusion Decoding (MRFD)}, a training-free decoding method that improves factual grounding by modeling inter-region consistency. MRFD identifies salient regions using cross-attention, generates initial responses for each, and computes reliability weights based on Jensen-Shannon Divergence (JSD) among the responses. These weights guide a consistency-aware fusion of per-region predictions, using region-aware prompts inspired by Chain-of-Thought reasoning. Experiments across multiple LVLMs and benchmarks show that MRFD significantly reduces hallucinations and improves response factuality without requiring model updates. Code is available at \href{https://github.com/Haonan-Ge/MRFD}{https://github.com/Haonan-Ge/MRFD}.
\end{abstract}

\begin{figure}[t]
    \centering
    \includegraphics[width=1\columnwidth]{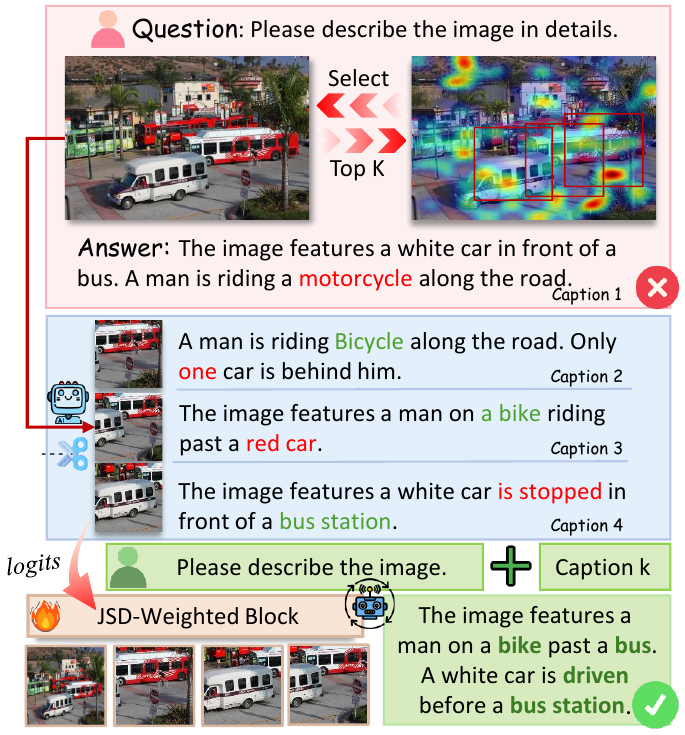} 
    \caption{The MRFD process: leveraging multiple regional responses (Captions 1-4), a JSD-Weighted Block derives consistency weights to guide a prompted fusion decoding, yielding a more reliable output.}
    \vspace{-4mm}
    \label{fig:intro}
\end{figure}

\section{Introduction}
\label{sec:introduction}

Large Vision-Language Models (LVLMs) have emerged as powerful tools for multimodal understanding, achieving significant advances in image captioning, visual question answering, and visual reasoning ~\cite{alayrac2022flamingo, liu2023visual, li2023blip2, zhu2023minigpt4,xia2024languagemultimodalmodelssports, li2025taco, li2025m2ivefficientfinegrainedmultimodal}. However, these models frequently generate hallucinated content, producing textual outputs inconsistent with the visual input. This phenomenon poses substantial challenges for reliability and practical deployment~\cite{ji2023survey, liu2024survey}. Hallucinations typically manifest as misidentified objects, fabricated attributes, or omitted visual information, reducing the factual alignment between image and text.

This issue is especially prominent in scenarios that require fine-grained understanding or interpretation of visually complex scenes. Current approaches include training-based solutions like data augmentation and fine-tuning, which improve factual grounding but require considerable resources and often lack generalization ~\cite{rohrbach2018object_chair, gunjal2024mitigating}. Training-free alternatives such as chain-of-thought prompting~\cite{wei2022chain} and contrastive or corrective decoding methods~\cite{li2023contrastive, leng2024mitigating_vcd, favero2024mitigating_m3id, woo2024ritual, huang2024opera, zhang2025selfcorr} offer flexibility, but they often rely on carefully crafted prompts or fail to exploit intrinsic visual information fully.

A key limitation shared by many of these approaches is their tendency to process images holistically or to analyze regions in isolation, often without dynamic mechanisms to assess the reliability of different visual cues or to reconcile potentially conflicting interpretations from multiple perspectives. Consequently, explicit consistency-based reasoning across various image segments is rarely integrated into current decoding procedures~\cite{feng2024details,zhang2021consensus}.

Our analysis (Section~\ref{sec:motivation}) highlights that some image regions provide more trustworthy evidence than others. In particular, we observe that the consistency of region-level responses, which are quantified by Jensen-Shannon Divergence (JSD), correlates with their factual correctness (see Section~\ref{subsec:Region_consistency}). This strong correlation indicates that inter-region agreement is a key determinant of output quality: responses consistent with the consensus view from multiple regions are not only more factually accurate but also demonstrate higher reliability and lower hallucination rates.

To this end, we propose \textbf{Multi-Region Fusion Decoding (MRFD)}, a decoding strategy that aims to incorporate multi-perspective reasoning into the generation process. As illustrated in Figure~\ref{fig:intro}, MRFD identifies several salient regions in the image based on attention mechanisms, then generates an initial response for each. By computing Jensen-Shannon Divergence (JSD) among these responses, the method derives a set of consistency-based reliability weights. These weights are used during decoding to combine predictions from all regions. Furthermore, inspired by CoT prompting, we construct region-aware prompts by combining the original question with the initial analysis of each region, preserving the localized context throughout the generation. The contributions of this work are:
\begin{itemize}
    \item We propose a training-free decoding method that integrates multiple region-level perspectives, weighted by inter-region consistency, to reduce hallucinations in LVLMs.
    \item We introduce a JSD-based scoring scheme to quantify agreement among region-wise responses and guide reliability-aware fusion.
    \item We design a region-aware prompting strategy to enhance contextual grounding during generation without modifying model parameters.
\end{itemize}

\section{Related Work}
\textbf{LVLM Hallucinations and Grounding Deficiencies.}
Large Vision-Language Models (LVLMs) frequently 'hallucinate'—generating text unsupported by visual input~\cite{ji2023survey, liu2024survey}. This often stems from flawed visual grounding due to various factors such as data biases or poor connections between vision and language components~\cite{han2024analyzing, jiang2024revealing, tong2024misalignment, zhou2024identifying}. Although standard attention mechanisms~\cite{anderson2018bottom} and region-based analyzes~\cite{li2022grounded, kamath2021mdetr} aim to improve grounding, LVLMs still struggle to reliably combine information from multiple, potentially conflicting, image regions or to assess their mutual consistency.

\medskip
\noindent\textbf{Limitations in Advanced Decoding and Fusion Strategies.}
Advanced strategies to improve LVLM outputs also exhibit limitations. \textit{Chain-of-Thought} (CoT) prompting~\cite{wei2022chain, zhang2023multimodal, lyu2023faithful}, while enhancing reasoning, may not ensure its steps visually align with the image and can be sensitive to setup or resource-intensive. Many training-free corrective or contrastive decoding methods (e.g.,~\cite{li2023contrastive, leng2024mitigating_vcd, favero2024mitigating_m3id, woo2024ritual, huang2024opera, zhang2025selfcorr, wang2024mitigating_disturbance}) struggle to capture finegrained local image features and risk missing valid information, resulting in poor grounding capabilities. Moreover, when attempting to fuse information from multiple sources—a concept for which consistency has proven beneficial in language modeling tasks~\cite{wang2023self_consistency, xiong2023can}—LVLMs face distinct challenges with visual regions. Common simplistic aggregation techniques (like averaging) are often insufficient for visual data, where the trustworthiness of different regions can vary dramatically. A central challenge thus remains: developing principled methods to appropriately weigh and fuse evidence from diverse, contextually-understood visual regions based on their consistency.

\section{Motivation}
\label{sec:motivation}

\begin{figure}[t]
    \centering
    \includegraphics[width=1\columnwidth]{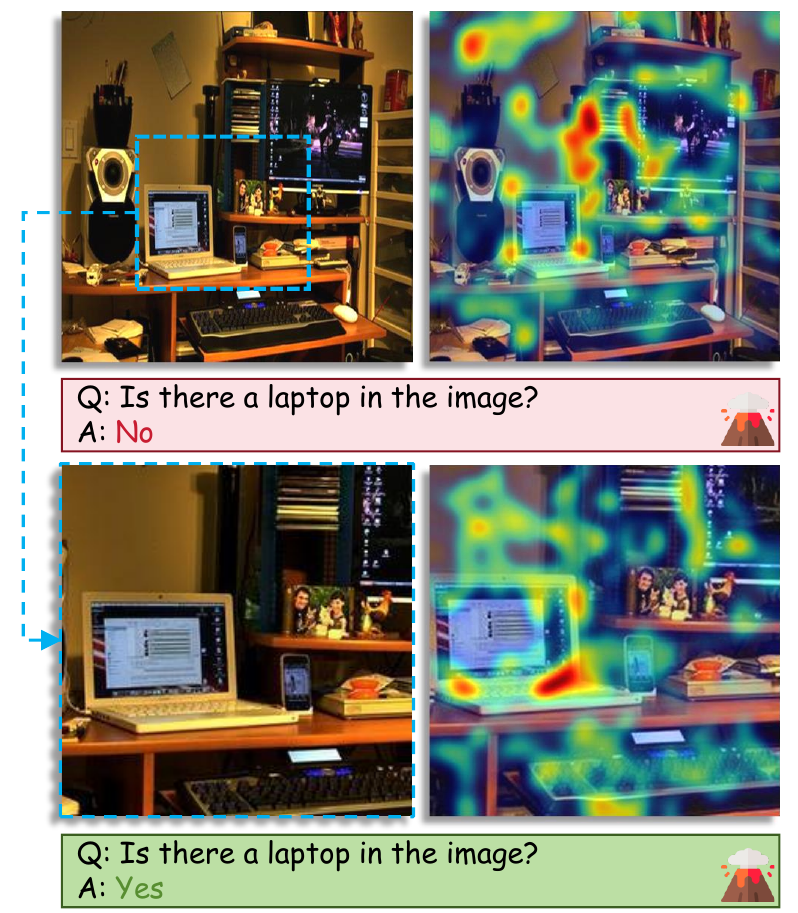} 
    \caption{LVLM cross-attention patterns for "Is there a laptop in the image?". (\textit{upper}) Full image input results in scattered attention and potential error. (\textit{lower}) Cropped image input focused on the laptop yields concentrated attention and improved accuracy.}
    \vspace{-3mm}
    \label{fig:attention_patterns}
    
\end{figure}

\subsection{Global Decoding Misses Local Evidence}
\label{sec:global_decoding_motivation}

To explore whether visual grounding capability is related to the hallucinations in LVLMs, we analyze their attention patterns during question answering. Figure~\ref{fig:attention_patterns} shows a case where the model is asked: “Is there a laptop in the image?” When using the full image as input, the model outputs “NO,” even though a laptop is present. The attention map shows that the model distributes focus across unrelated regions.

In contrast, when the image is cropped to a salient region based on high attention (the desk area), the model concentrates more narrowly on the relevant evidence and correctly answers “YES.” The localized input guides the model to verify only what matters for the query, avoiding distractions from the rest of the image.

This comparison reveals a key shortcoming of global decoding: attention is easily diffused across many areas, some of which may mislead the model. In cluttered or ambiguous scenes, this often results in factual errors. Cropped regions help narrow the visual focus of the model, improving the quality of the answer by strengthening the localized foundation.

\begin{figure}[t]
    \centering
    \includegraphics[width=1\columnwidth]{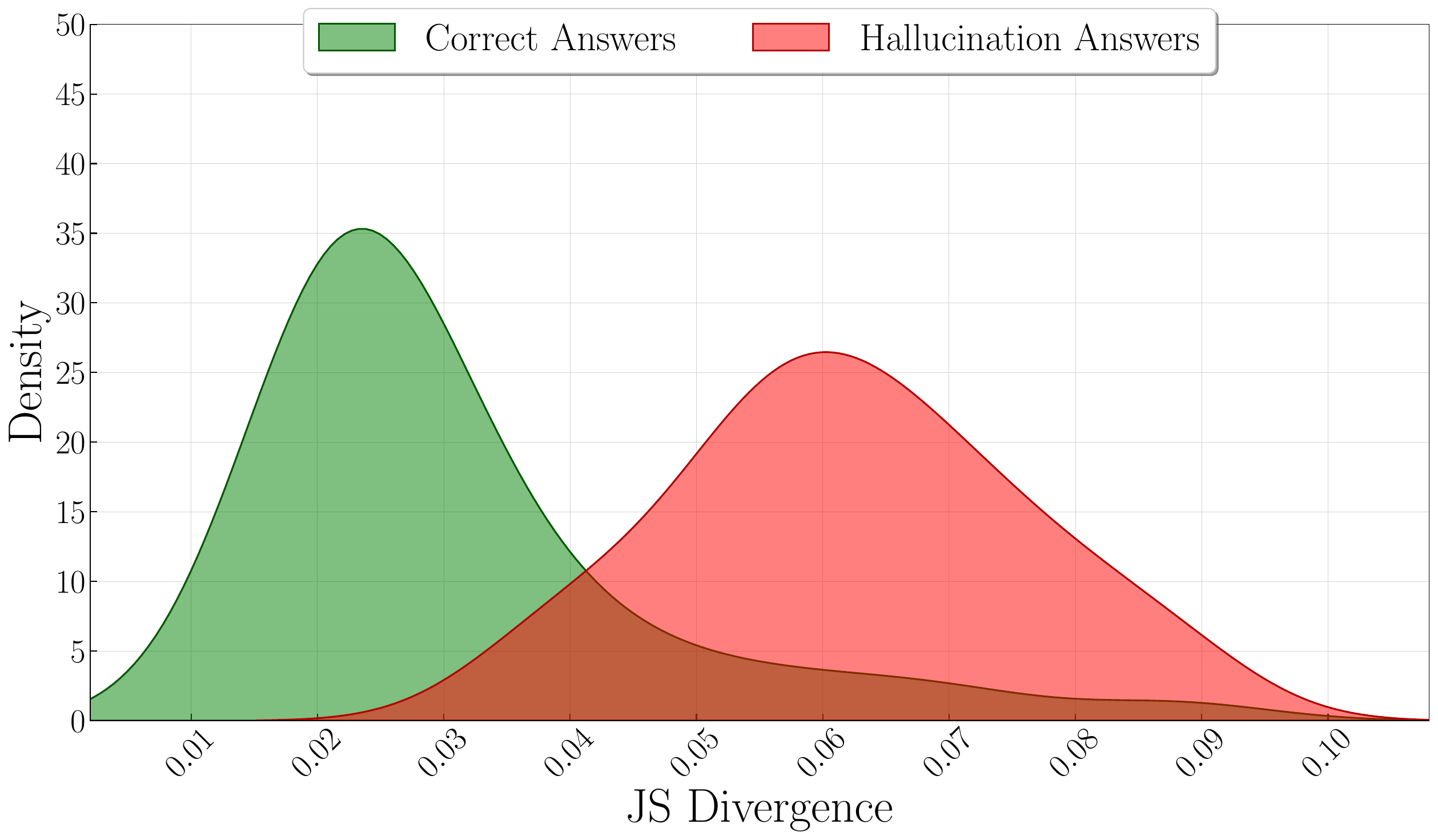} 
    \caption{Density distribution of JS Divergence for correct versus hallucinated LVLM responses, indicating lower JSD correlates with higher factual accuracy.}
    \label{fig:jsd_density_plot}
    \vspace{-3mm}
\end{figure}
\subsection{Region Consistency Reflects Reliability}
\label{subsec:Region_consistency}

However, not all regions—even attention-guided ones—are equally reliable. Some may be visually ambiguous or contextually misleading. To avoid overconfident on a single (possibly misleading) region, we adopt a self-consistency approach: comparing multiple region-level responses. When different views agree on an answer, that answer is more likely to be trustworthy.

To test this, we run experiments using LLaVA-1.5 on 3,000 MSCOCO validation samples, annotated with hallucination labels from the POPE benchmark. For each image-question pair, we generate several responses from attention-guided image patches, along with one from the full image. For the output of each cropped region, we calculate the Jensen-Shannon divergence (JSD) between its response distribution and the average response distribution across all regions.

Our results (illustrated in Figure~\ref{fig:jsd_density_plot}) show a clear distinction: JS Divergence for hallucinated answers tends to concentrate around 0.06-0.07, whereas for correct answers, it centers around approximately 0.02. This indicates that the JSD between a specific regional response and the average across regions effectively reflects the level of hallucination of that regional response: a lower JSD correlates with fewer hallucinations. This finding resonates with the principle of self-consistency~\cite{wang2023self_consistency}, where agreement among multiple diverse outputs often signals higher quality and reliability. Motivated by this, we design a decoding strategy: A mechanism that aggregates evidence from multiple regions, and checks their agreement, can help improve the robustness and factual grounding of model predictions.

\begin{figure*}[!tb]
\centering
\includegraphics[width=1.0\linewidth]{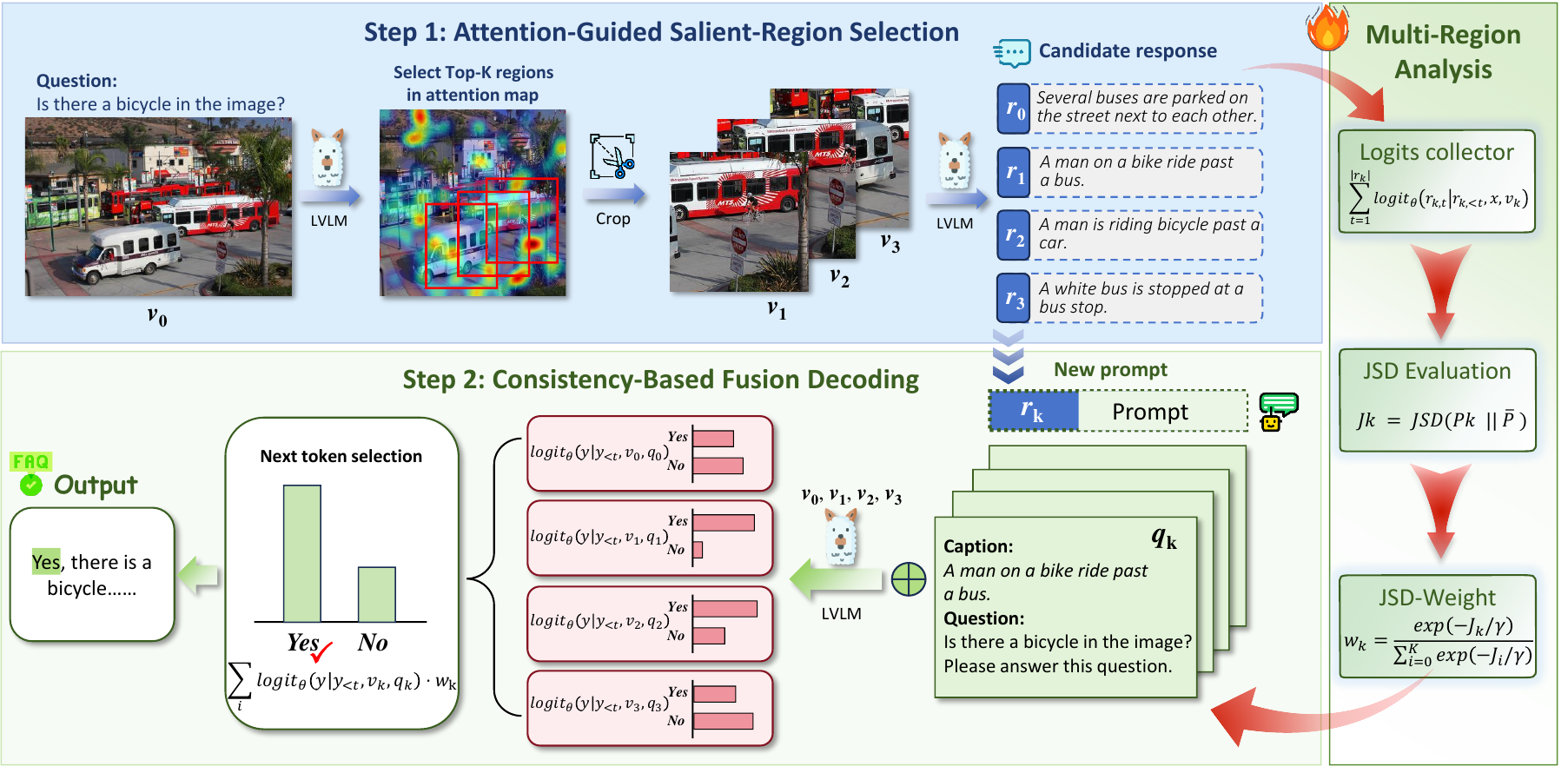} 
\caption{Overall framework of Multi-Region Fusion Decoding (MRFD): Step 1 uses 
attention to select and crop salient regions ($v_k$), generates candidate responses ($r_k$) per region, 
and computes JSD-based consistency weights ($w_k$) for each response. 
Step 2 forms new inputs per region with a candidate response and the original prompt. They are all processed in parallel,
 fusing per-region logits using the weights $w_k$  during parallel decoding to select the output tokens.}
\label{fig:method} 
\end{figure*}
\section{Method}

\subsection{Overview}

Current LVLM decoding methods often suffer from hallucinations due to the lack of multi-perspective consistency checking, as discussed in Section~\ref{sec:motivation}.

To address this, we propose \textbf{Multi-Region Fusion Decoding (MRFD)}, a training-free decoding strategy that enhances answer reliability through:
1) Attention-Guided Region Selection: selecting salient regions based on attention maps;
2) Multi-Region Analysis with JSD-Based Weighting: assessing consistency across regions using Jensen-Shannon Divergence;
3) Consistency-Based Fusion Decoding: fusing predictions weighted by their mutual consistency.
The full framework is illustrated in Figure~\ref{fig:method}.

\subsection{Attention-Guided Region Selection}
Building on the multi-perspective verification approach motivated in Section~\ref{sec:motivation}, the first crucial step in our MRFD’s framework is to identify and isolate multiple informative regions from the input image for focused and independent analysis.
\medskip

\noindent\textbf{Leveraging Cross-Attention for Region Identification.} To identify image regions most relevant to the input query, we leverage the cross-attention mechanisms already present in LVLM architectures. Given an original image $I_0$ and a text query $x$, we compute the attention matrix $A$ from the LVLM's cross-attention layers. Let $H_T \in \mathbb{R}^{n \times d}$ represent the sequence of hidden states from the textual input, and $H_V \in \mathbb{R}^{m \times d}$ represent those from the visual input, where $n$ and $m$ denote sequence lengths and $d$ represents the hidden dimensionality.

The cross-attention weight matrix $A \in \mathbb{R}^{n \times m}$, capturing alignment between textual and visual modalities, is computed using scaled dot-product attention\cite{vaswani2017attention}:
\begin{equation} \label{eq:attention} 
    \mathbf{A} = \text{softmax}\left(\frac{(H_{T} W^Q) (H_{V} W^K)^\top}{\sqrt{d_k}}\right)
    \end{equation}
    where $W^Q \in \mathbb{R}^{d \times d_k}$ and $W^K \in \mathbb{R}^{d \times d_k}$ are projection matrices for queries and keys, and $d_k$ is their dimensionality.

To obtain an overall map of visual focus, we aggregate these attention weights across the textual dimension:
\begin{equation} \label{eq:attention_aggregation}
    a_j = \frac{1}{n}\sum_{i=1}^{n}A_{ij} \quad \text{for\ } j = 1, \ldots, m
    \end{equation}
    This produces an aggregated attention vector $a \in \mathbb{R}^{1 \times m}$. Assuming the visual hidden states correspond to a spatial grid of patches ($d' \times d' = m$), we reshape this vector into a 2D spatial attention map:
\begin{equation} \label{eq:attention_reshape}
\hat{A} = \text{reshape}(\mathbf{a}, (d', d'))
\end{equation}
This spatial attention map $\hat{A}$ visually highlights the regions that the model finds most relevant to the input query. By utilizing the model's own attention patterns, we can directly identify which parts of the image are being primarily considered when answering the query.

\medskip
\noindent\textbf{Selecting Salient Regions. }Using the spatial attention map $\hat{A}$, we identify the top-K most salient regions ${R_1, R_2, ..., R_K}$ within the original image. To ensure diversity and comprehensive coverage, we employ an integral image approach to efficiently search for K non-overlapping or minimally overlapping regions that maximize summed attention scores. For practical implementation, we constrain these regions to be rectangular bounding boxes with a maximum overlap (IoU) threshold between any two regions to ensure diversity in the selected perspectives.

For each selected region $R_k$, we crop the original image to obtain a focused sub-image $v_k$. This process yields a set of K+1 visual inputs: the original full image $v_0 = I_0$ and K cropped region images $\{v_1, v_2, ..., v_K\}$. These cropped regions represent the most informative parts of the image with respect to the input query, as determined by the model's own attention mechanism.


\subsection{Multi-Region Analysis with JSD-Based Weighting}

Having selected salient regions, we now detail their independent analysis and our JSD-based weighting mechanism. This approach is crucial for assessing inter-region consistency(Section~\ref{subsec:Region_consistency}).
For each selected region $R_k$, we process the corresponding cropped sub-image $v_k$ through the LVLM to generate an initial analysis:
\begin{equation}\label{eq:region_analysis}
r_k = \text{LVLM}(v_k, x)
\end{equation}
where $r_k$ represents the LVLM's response when considering only the visual information from region $k$. We also generate a response $r_0$ using the original full image $v_0$. The resulting set $\{r_0, r_1, ..., r_K\}$ provides multiple perspectives on the query, each grounded in different salient regions of the image.


To assess the reliability of information from each region, we measure the consistency between region-specific analyses using Jensen-Shannon Divergence (JSD). The JSD provides a symmetric measure of similarity between probability distributions and is defined as:
\begin{equation} \label{eq:jsd}
    JSD(P || Q) = \frac{1}{2} D_{KL}(P || M) + \frac{1}{2} D_{KL}(Q || M)
    \end{equation}
    where $P$ and $Q$ are two probability distributions, $M = \frac{1}{2}(P + Q)$, and $D_{KL}$ is the Kullback-Leibler divergence.

We derive a representative vocabulary distribution $P_k$ for each initial analysis sequence $r_k$ by averaging the next-token probability distributions computed during generation:
\begin{equation} \label{eq:probability_distribution}
    P_k = \text{softmax}(\frac{1}{|r_k|} \sum_{t=1}^{|r_k|} \text{logit}_{\theta}(y|r_{k,<t}, x, v_k))
    \end{equation}
    where $|r_k|$ is the length of $r_k$ and $\text{logit}_\theta$ represents the model's output logits. This distribution $P_k$ captures the overall token prediction patterns when the model considers region $k$.
We then compute the average distribution across all regions:
\begin{equation} \label{eq:p-average}
    \overline{P} = \frac{1}{K+1} \sum_{i=0}^{K} P_i
    \end{equation}
For each region $k$, we calculate its divergence from this average distribution:
\begin{equation} \label{eq:jsd_score}
J_k = JSD(P_k||\bar{P})    
\end{equation}
These divergence scores measure how much each region's predictions deviate from the consensus. A lower $J_k$ indicates that region $k$ provides information more consistent with other regions, suggesting higher reliability.

Finally, we convert these divergence scores into weights with temperature parameter $\gamma$:
\begin{equation} \label{eq:weight}
    w_k = \frac{\exp(-J_k / \gamma)}{\sum_{i=0}^{K} \exp(- J_i / \gamma)}
    \end{equation}
    The temperature parameter $\gamma$ controls the sharpness of the weight distribution—a smaller $\gamma$ creates more contrast between weights, while a larger $\gamma$ leads to more uniform weighting.


\subsection{JS-Weighted Integrative Fusion Decoding}
With the consistency-based weights $w_k$ computed for each visual input $v_k$ (k = 0...K), we now perform the final decoding step by fusing the next-token predictions in a manner that prioritizes consistent visual evidence.

Motivated by the effectiveness of Chain-of-Thought (CoT) prompting~\cite{wei2022chain} in structuring reasoning, we adapt its core principle to enhance factual grounding within our multi-region fusion framework. For each region $k$, we construct a fixed region-aware prompt $q_k$ by concatenating the original question $x$ and the corresponding region's initial analysis $r_k$:
\begin{equation}
    q_k = \text{Concat}(x, r_k)
    \label{eq:region_aware_prompt}
    \end{equation}
    This prompt $q_k$ encapsulates both the global query and the localized preliminary analysis derived from $v_k$. It serves as a static, enriched context for region $k$ throughout the decoding process, providing region-specific grounding information without requiring dynamic prompt updates.

During autoregressive decoding, at each step $t'$, the LVLM produces next-token logits $\ell_k^{(t')}$ conditioned on the current partial output sequence $y_{<t'}$, the visual input $v_k$, and the fixed region-aware prompt $q_k$:
\begin{equation}
    \ell_k^{(t')} = \text{logit}_\theta(y | y_{<t'}, v_k, q_k)
    \end{equation}
    These logits represent the prediction from each region's perspective given its pre-computed analysis. We aggregate them using the pre-computed consistency weights $w_k$:
    \begin{equation}
        \ell_{\text{fused}}^{(t')} = \sum_{k=0}^{K} w_k \cdot \ell_k^{(t')}
        \label{eq:weighted_fusion}
        \end{equation}

The final probability distribution for the next token is obtained via softmax:
\begin{equation}
    P_{\text{fused}}^{(t')}(y) = \text{softmax}(\ell_{\text{fused}}^{(t')})
    \end{equation}
    The next token $\hat{y}_{t'}$ is then selected (e.g., sampling) from this fused distribution, and the process repeats autoregressively until completion.

    

\section{Experiments}
\label{sec:experiments}
\subsection{Experimental Settings}
\noindent\textbf{Evaluated LVLMs.}
We evaluate Multi-Region Fusion Decoding (MRFD) on two representative open-source LVLMs: LLaVA-1.5-7B~\cite{liu2023improvedllava} and InstructBLIP-7B~\cite{dai2023instructblip}. These models adopt different vision-language interfaces—direct projection in LLaVA-1.5 and query-based encoding (Q-Former~\cite{li2023blip2}) in InstructBLIP—enabling evaluation of MRFD’s generality. MRFD is applied as a training-free, decoding-time procedure on frozen models. Architectural details are provided in Appendix~\ref{sec:appendix_evaluated_lvms}.

\medskip

\noindent\textbf{Baselines.}
As a basic baseline, regular decoding samples tokens from the model's post-softmax output probabilities. These include contrastive methods such as VCD~\cite{leng2024mitigating_vcd} and M3ID~\cite{favero2024mitigating_m3id}, robustness-enhancing methods like RITUAL~\cite{woo2024ritual}, and approaches employing correction or feedback, such as DeGF~\cite{zhang2025degf} and Woodpecker~\cite{yin2023woodpecker}. We also include comparisons with other relevant methods: HALC~\cite{chen2024halc}, and OPERA~\cite{huang2024opera}. Performance of these baselines is based on our re-implementations using publicly available code where possible. Detailed descriptions of each baseline's methodology are provided in Appendix~\ref{sec:appendix_baseline_details}.

\medskip

\noindent\textbf{Implementation Details.} 
Across all experiments, our Multi-Region Fusion Decoding (MRFD) method selects $K=3$ salient regions and employs a temperature of $\gamma=0.02$ for the JSD-based weighting (Eq.~\ref{eq:weight}). We utilize multinomial sampling for both stages of decoding. Other detailed settings are provided in Appendix~\ref{sec:appendix_implementation_details}.

\subsection{Datasets and Benchmarks}

We evaluate MRFD on multiple benchmarks covering both hallucination detection and general vision-language understanding. Below we briefly describe the key datasets; full details and evaluation metrics are provided in Appendix~\ref{sec:data}.

\noindent\textbf{POPE}~\cite{li2023evaluating_pope}: A Yes/No QA benchmark for object existence hallucination, built from MSCOCO, A-OKVQA, and GQA with various negative sampling strategies.

\noindent\textbf{CHAIR}~\cite{rohrbach2018object_chair}: Measures hallucinated object mentions in image captions via CHAIRi and CHAIRs scores on MSCOCO images.

\noindent\textbf{MME}~\cite{fu2023mme}: We focus on the MME-Hallucination subset, which tests object, count, position, and attribute hallucinations.


\subsection{Results and Discussions}

\begin{table*}[t]
    
    \centering
    \small
    \resizebox{\textwidth}{!}{%
    \begin{tabular}{lllcccccc}
    \toprule
    \multirow{2}{*}{\textbf{Setting}} & \multirow{2}{*}{\textbf{Method}} & & \multicolumn{3}{c}{\textbf{LLaVA-1.5}\llava} & \multicolumn{3}{c}{\textbf{InstructBLIP}\insblip} \\
    \cmidrule(lr){4-6} \cmidrule(lr){7-9}
    & & & Acc. $\uparrow$ & Prec. $\uparrow$ & F1 $\uparrow$ & Acc. $\uparrow$ & Prec. $\uparrow$ & F1 $\uparrow$ \\
    \midrule
    \multirow{6}{*}{\textit{Random}} & Regular & & 82.42 & 78.30 & 83.67 & 79.85          & 80.33          & 83.45          \\
    & VCD\vcd     & & 84.69 & 80.30 & 85.85 & 84.47          & 83.07          & 84.81          \\
    & M3ID\md    & & 85.46 & 81.54 & 86.42 & 85.32          & 83.51          & 85.58          \\
    & RITUAL\ritual  & & 86.71 & 82.84 & 87.51 & 87.12          & 85.64          & \underline{87.23} \\
    & DeGF\degf    & & \underline{87.79} & \underline{86.33} & \underline{88.08} & \underline{87.21} & \textbf{89.01} & 86.70          \\
    \rowcolor{green!26}  \cellcolor{white}& \textbf{MRFD}    & & \textbf{88.15} & \textbf{88.91} & \textbf{88.23} & \textbf{88.03} & \underline{88.74} & \textbf{88.01} \\
    \midrule
    \multirow{6}{*}{\textit{Popular}} & Regular & & 76.57 & 71.23 & 79.56 & 75.17          & 70.90          & 77.54          \\
    & VCD\vcd     & & 77.30 & 71.61 & 80.57 & 78.12          & 73.77          & 80.10          \\
    & M3ID\md    & & 78.66 & 73.09 & 81.45 & 78.32          & 73.75          & 80.30          \\
    & RITUAL\ritual  & & 79.75 & 74.55 & 82.31 & 78.40          & 73.63          & 80.55          \\
    & DeGF\degf    & & \underline{81.94} & \underline{78.33} & \underline{83.31} & \underline{80.37} & \underline{78.84} & \underline{81.24} \\
    \rowcolor{green!26}  \cellcolor{white}& \textbf{MRFD}     & & \textbf{81.99} & \textbf{78.81} & \textbf{83.29} & \textbf{82.17} & \textbf{79.52} & \textbf{83.15} \\
    \midrule
    \multirow{6}{*}{\textit{Adversarial}} & Regular & & 71.09 & 65.77 & 75.93 & 71.02          & 66.58          & 74.70          \\
    & VCD\vcd     & & 71.13 & 65.28 & 76.37 & 73.07          & 68.50          & 76.36          \\
    & M3ID\md    & & 72.10 & 66.27 & 76.87 & 72.97          & 67.92          & 76.55          \\
    & RITUAL\ritual  & & 71.87 & 66.17 & 76.88 & 73.06          & 67.83          & 76.91          \\
    & DeGF\degf    & & \underline{76.13} & \underline{71.50} & \underline{79.01} & \underline{75.96} & \underline{73.45} & \underline{77.36} \\
    \rowcolor{green!26}  \cellcolor{white}& \textbf{MRFD}     & & \textbf{77.99} & \textbf{76.16} & \textbf{79.22} & \textbf{77.72} & \textbf{74.26} & \textbf{79.72} \\
    \bottomrule
    \end{tabular}%
    }
    \caption{Results on POPE benchmark. Higher ($\uparrow$) accuracy, precision, and F1 indicate better performance. The best results are bolded, and the second-best are underlined.}
    \label{tab:pope_results}
    \end{table*}

\noindent\textbf{Results on POPE.}
In Table~\ref{tab:pope_results}, we compare the performance of our MRFD method against other baselines on the POPE benchmark~\cite{li2023evaluating_pope} under three different negative sampling settings (Random, Popular, Adversarial), across both LLaVA-1.5 and InstructBLIP. As shown, MRFD consistently outperforms other decoding methods on both LVLMs, achieving leading F1 scores across all six configurations, with improvements of up to 2.44\% in accuracy, 6.52\% in precision, and 3.05\% in F1 score compared to the respective second-best approaches. This suggests that MRFD's core strategy—identifying multiple salient regions via cross-attention, assessing their response consistency using JSD-weighting, and fusing their context-enriched predictions—enables LVLMs to better ground responses in relevant visual evidence, thereby effectively addressing object hallucinations. Moreover, while most decoding method tend to be overcofident in nonexistence, the consistency verification inherent in MRFD appears to promote more cautious and precise responses, which is evidenced by its strong precision, particularly in challenging adversarial settings (e.g., 76.16 on LLaVA-1.5 and 74.26 on InstructBLIP), highlighting its enhanced capability in filtering false positives and suppressing misinformation. Detailed results of POPE are attached in Appendix~\ref{detailed_pope}.

\begin{table}[t]
    \centering
    \scriptsize
    \resizebox{\columnwidth}{!}{%
    \begin{tabular}{lccccc}
    \toprule
    \multirow{2}{*}{{Method}} & \multicolumn{2}{c}{{LLaVA-1.5}} & \multicolumn{2}{c}{{InstructBLIP}} \\
    \cmidrule(lr){2-3} \cmidrule(lr){4-5}
    & Cs $\downarrow$ & Ci $\downarrow$ & Cs $\downarrow$ & Ci $\downarrow$ \\
    \midrule
    Regular    & 26.2          & 9.4           & 31.2          & 11.1 \\
    VCD        & 24.4          & 7.9           & 30.0          & 10.1 \\
    M3ID       & 21.4 & 6.3 & 30.8          & 10.4 \\
    RITUAL     & 22.4          & 6.9           & 26.6          & 8.9  \\
    Woodpecker & 24.9          & 7.5           & 31.2          & 10.8 \\
    HALC       & 21.7          & 7.1           & 24.5 & 8.0 \\
    DeGF       & \underline{18.4}          & \underline{6.1}           & \underline{24.0}          & \underline{7.7}  \\
    \midrule
    \rowcolor{green!26} \textbf{Ours (MRFD)} & \textbf{14.1} & \textbf{5.0}  & \textbf{21.3} & \textbf{6.1} \\
    \bottomrule
    \end{tabular}
    }
    \caption{Results on CHAIR benchmark for caption generation. We limited the maximum number of new tokens to 64. Lower ($\downarrow$) CHAIRs (Cs) and CHAIRi (Ci) indicate less hallucination. Best results are bolded, second-best are underlined.}
    \vspace{-2mm}
    \label{tab:chair_results}
\end{table}

\medskip
\noindent\textbf{Results on CHAIR.}
We evaluate MRFD's effectiveness in mitigating object hallucination in open-ended image captioning using the CHAIR benchmark~\cite{rohrbach2018object_chair}, reporting CHAIRs (Cs) and CHAIRi (Ci) scores (lower is better) for LLaVA-1.5 and InstructBLIP in Table~\ref{tab:chair_results}. MRFD consistently achieves state-of-the-art performance on both LVLMs, significantly outperforming the strong DeGF baseline with relative CHAIR score reductions of up to 16.4\% on LLaVA-1.5 and 20.8\% on InstructBLIP. Given that the CHAIR task for image captioning necessitates comprehensive attention to diverse local details, MRFD's superior performance underscores its advanced capability for robust multi-region analysis and effective integration of key information from various visual segments, leading to more factually grounded image captions. Detialed results of CHAIR are appended to Appendix~\ref{detailed_chair}. Some qualitative examples are shown in Appendix~\ref{sec:appendix_qualitative_chair}.

\begin{figure}[t]
    \centering
    \includegraphics[width=1\columnwidth]{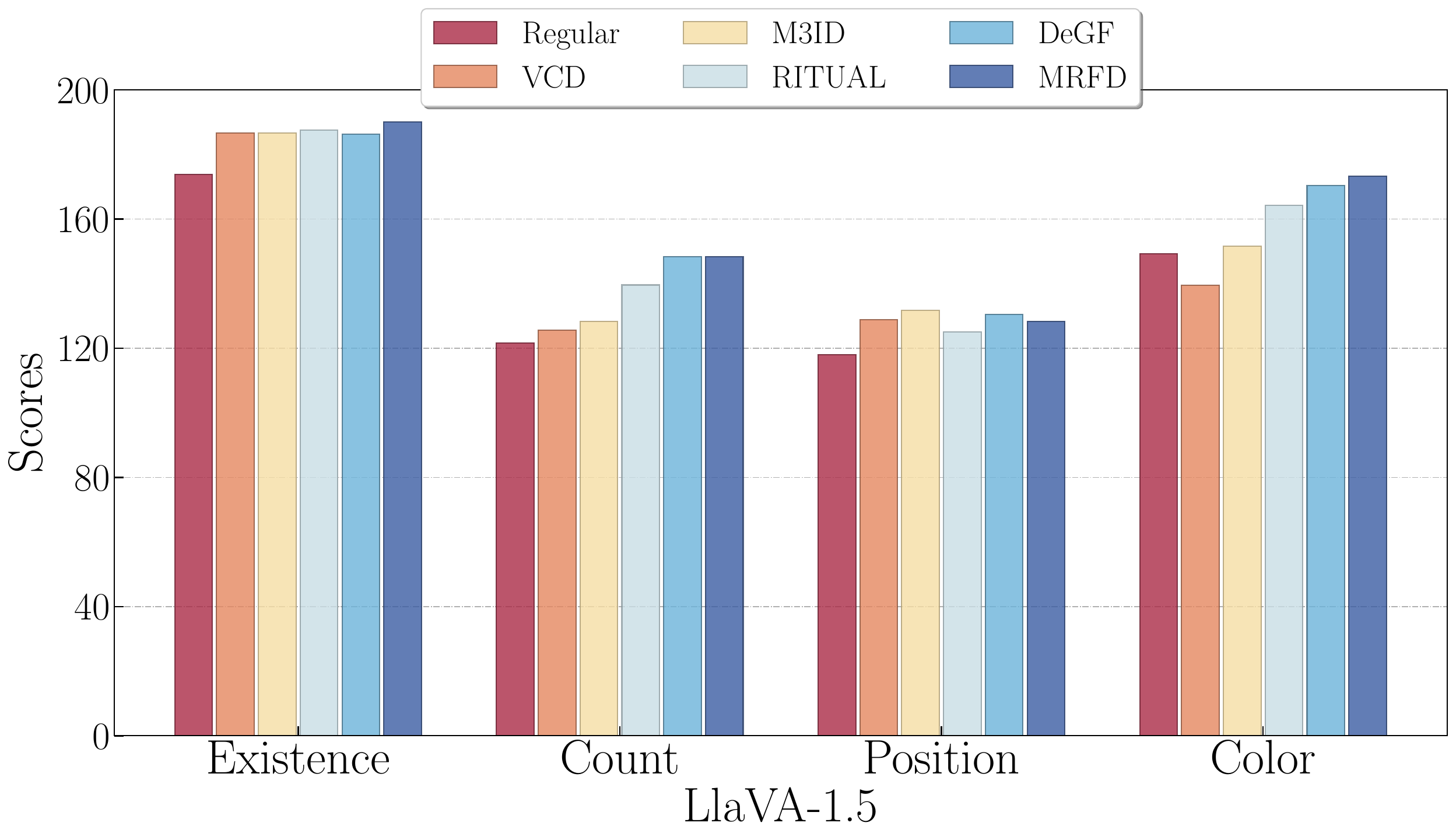} 
    \caption{Experimental results of MME on a hallucination subset with different decoding strategies.}
    \label{fig:mme}
\end{figure}

\medskip
\noindent\textbf{Results on MME-Hallucination.}
We evaluate MRFD on the MME hallucination subset~\cite{fu2023mme}, which assesses diverse hallucination types including object-level (Existence), attribute-level (Count, Color), and relation-level (Position) understanding in LVLMs. As illustrated in Figure~\ref{fig:mme} (presenting results for both LLaVA-1.5 and InstructBLIP), our MRFD method consistently demonstrates strong performance, achieving leading or state-of-the-art results across the majority of these hallucination categories for both evaluated LVLMs when compared to regular decoding and other advanced baselines. This robust performance across a spectrum of challenging hallucination types underscores that MRFD's core mechanism—multi-region analysis coupled with consistency-driven fusion—effectively enhances fine-grained visual understanding and overall factual consistency. Detialed results of MME-Hallucination are appended to Appendix~\ref{detailed_mme}

\medskip
\noindent\textbf{Efficiency Discussion.}
Despite employing multiple inference steps, MRFD maintains a competitive efficiency profile, particularly when compared to other advanced hallucination mitigation techniques that involve more complex iterative or feedback mechanisms. Details are in Appendix~\ref{detailed_efficiency}.

\subsection{Ablation Study}
\label{sec:ablation}
We conduct ablation studies to evaluate the impact of key components and hyperparameters in MRFD, using LLaVA-1.5 on the POPE-MSCOCO dataset.

First, we assess the contributions of MRFD's core designs: JSD-based Consistency Weighting (CW), Fusion Prompt (FP), and attention-guided Region Selection (RS). We test our full MRFD against three main variants where these components are individually altered: (1) \textbf{MRFD w/o CW}, applying uniform fusion weights; (2) \textbf{MRFD w/o FP}, using only the original question for regional decoding; and (3) \textbf{MRFD w/o RS}, processing only the global image through the subsequent pipeline stages. As detailed in Table~\ref{tab:ablation_components_pope}, full MRFD achieves an 86.21 F1 score. Removing CW degrades F1 performance by 2.94\%, underscoring the importance of dynamic, consistency-based weighting. Omitting FP reduces F1 by 3.93\%, highlighting the value of enriched regional context. Bypassing RS results in the largest F1 drop of 4.07\%, emphasizing that robust multi-region analysis is fundamental. Despite these impacts, all three ablated MRFD variants still outperform Regular decoding (81.59 F1), while the complete MRFD configuration showcases the strongest synergistic benefits.

\begin{table}[t]
    \centering
    \begin{tabular}{lccc}
    \toprule
    Model Variants & Acc. $\uparrow$ & Prec. $\uparrow$ & F1 Score $\uparrow$ \\
    \midrule
    \textbf{MRFD (Full)} & \textbf{86.50} & \textbf{88.11} & \textbf{86.21} \\
    \midrule
    w/o CW & 83.76 & 84.27 & 83.74 \\ 
    w/o FP & 82.87 & 83.95 & 82.70 \\ 
    w/o RS & 82.77 & 83.94 & 82.58 \\ 
    Regular & 80.57 & 78.84 & 81.59 \\
    \bottomrule
    \end{tabular}
    \caption{Ablation study with different model variants on POPE-COCO under the average of three settings .}
    \label{tab:ablation_components_pope}
    \end{table}

\begin{figure}[t]
    \centering
    \includegraphics[width=1\columnwidth]{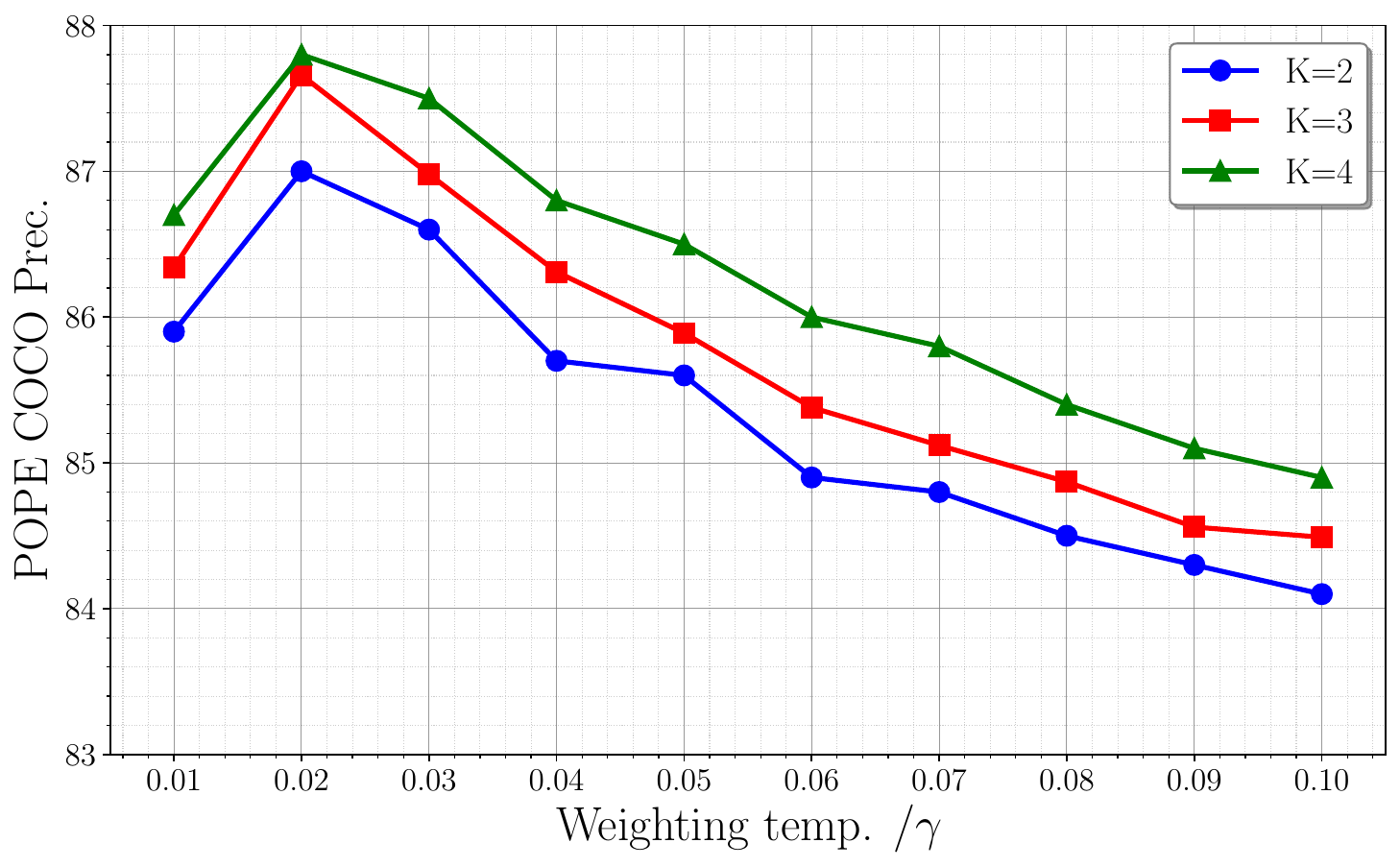} 
    \caption{Sensitivity of POPE-COCO Precision (LLaVA-1.5) to JSD weighting temperature $\gamma$ for $K=2,3,4$ regions, with optimal performance around $\gamma=0.02$.}
    \label{fig:gamma_sensitivity_pope}
    \vspace{-3mm}
\end{figure}
Second, we analyze the sensitivity to the JSD weighting temperature $\gamma$ (Eq.~\ref{eq:weight}). As depicted in Figure~\ref{fig:gamma_sensitivity_pope} (POPE COCO Precision for $K=2,3,4$), performance peaks around $\gamma=0.02$ ($K=3$ at $\approx$87.8\% Prec.). Higher $\gamma$ values ($\ge$0.08) lead to more uniform weights and correspondingly reduced precision, approaching unweighted fusion. Conversely, very low $\gamma$ values ($\approx$0.01) cause over-reliance on a few regions due to extreme weighting, which can negatively impact robustness, particularly for larger $K$ (e.g., performance degradation for $K=4$). Thus, $\gamma=0.02$ is adopted for providing an optimal balance. Further details are presented in Appendix~\ref{detailed_ablation}.

\section{Conclusion}
We presented Multi-Region Fusion Decoding (MRFD), a training-free approach that improves LVLM reliability by mimicking self-consistency multi-view verification. MRFD identifies salient regions via attention, estimates their reliability using Jensen-Shannon Divergence (JSD), and fuses region-level predictions using consistency-weighted, Chain-of-Thought-inspired prompts. Experiments show that MRFD effectively reduces hallucinations and improves factuality across various LVLMs and benchmarks, offering a simple yet robust decoding strategy.

\section*{Limitations}
MRFD relies on the quality and interpretability of attention maps produced by the underlying LVLM, which may vary across architectures and input conditions. The assumption that inter-region consistency indicates factual correctness holds empirically but may not generalize to all reasoning tasks, especially those requiring abstract or commonsense inference. Additionally, while MRFD improves factuality in image-grounded tasks, it has not been evaluated in broader multimodal contexts such as video or dialogue-based grounding, which may involve more complex temporal or conversational dependencies.

\section*{Ethics Statement}
Our work on Multi-Region Fusion Decoding (MRFD) aims to improve the reliability and factual grounding of LVLMs, thereby contributing to more trustworthy AI systems by reducing visual hallucinations. We believe this has positive implications for applications where factual accuracy is critical.

\if
\section*{Acknowledgements}
The work is partially supported by the NSF of the United States Grant CRII 2451683, an NVIDIA Academic Grants Program, University of California at Merced, and a UC Merced Faculty Research Award.
The views and conclusions are those of the authors and should not reflect the official policy or position of the U.S. Government.

We also thank Prof. Kai-Wei Chang (Department of Computer Science, UCLA) for his insightful feedback on earlier drafts, including suggestions on clarifying terminology, adding appropriate cognitive-science references, discussing experimental settings, and presenting qualitative examples, as well as for pointing us to relevant related work. Any remaining errors are our own.
\fi

\bibliography{latex/custom}

\newpage
\appendix

\section{Evaluated LVLMs}
\label{sec:appendix_evaluated_lvms}
We evaluate our proposed Multi-Region Fusion Decoding (MRFD) framework on three representative open-source Large Vision-Language Models (VLMs): \textbf{LLaVA-1.5-7B} \cite{liu2023improvedllava}, and \textbf{InstructBLIP-7B} \cite{dai2023instructblip}. For the visual encoder, LLaVA-1.5 uses ViT-L-336px pre-trained from CLIP-L/14-336px \cite{radford2021learning}. InstructBLIP uses ViT-g/14 pre-trained from EVA-CLIP \cite{sun2023eva_clip}. Qwen-VL uses Openclip ViT-bigG \cite{Ilharco_OpenCLIP_2021_zenodo}. Regarding the language model module, LLaVA-1.5 and InstructBLIP both utilize Vicuna-7B \cite{chiang2023vicuna}.

The vision-language interface varies across the evaluated models. LLaVA-1.5 directly projects visual features using MLP layers. InstructBLIP, conversely, adopts the Q-Former \cite{li2023blip2}, which processes visual features via 32 learnable queries to generate a standardized input for the LLM. Our evaluation of MRFD encompasses these distinct strategies—direct MLP projection and query-based bottleneck (Q-Former)—to demonstrate its broad effectiveness in mitigating hallucinations regardless of the connection module employed.

\section{Baselines}
\label{sec:appendix_baseline_details}

We compare the performance of our MRFD method with several recent training-free decoding approaches designed to mitigate hallucination. These key baselines, along with their core methodologies, are described below:

\begin{itemize}
    \item \textbf{VCD}~\cite{leng2024mitigating_vcd}: This method contrasts output distributions derived from an original visual input ($v$) and a distorted version ($v'$). Given a textual query and $v$, the model generates two distributions. The distorted input $v'$ is created using pre-defined distortions (e.g., a Gaussian noise mask) to $v$. These two distributions are then contrasted to form the final output probability. For reproduction, we follow VCD's default setting with $\alpha = 1$ (a parameter controlling the contrast strength) and use 500 noise steps to generate $v'$. VCD aims to enhance visual grounding by reducing reliance on language priors.

    \item \textbf{M3ID}~\cite{favero2024mitigating_m3id}: M3ID contrasts output distributions from original visual inputs against those from pure text inputs (lacking visual information). The final probability distribution is a combination of the distribution conditioned on both vision and text, and a contrastive term derived from the difference between vision-conditioned and text-only conditioned distributions, balanced by a hyperparameter $\lambda$. We adhere to their recommended $\lambda = 0.02$. M3ID also seeks to improve visual grounding by emphasizing visual information.

    \item \textbf{RITUAL}~\cite{woo2024ritual}: RITUAL applies common image transformations (e.g., crop, flip, color jitter) to the original visual input $v$, creating a transformed version $v^{(T)}$. It then generates the response by utilizing information from both the original ($v$) and transformed ($v^{(T)}$) images. The final probability distribution combines logits from both views, with the contribution of the transformed input adjusted by a balancing hyperparameter $\kappa$. We follow their official implementation, setting $\kappa = 3$. This approach aims to improve robustness through consistency across augmented views.

    \item \textbf{DeGF}~\cite{zhang2025degf}: DeGF introduces a self-correction mechanism using feedback from text-to-image generative models. Specifically, it first generates an image based on the LVLM's initial textual response. This generated image then acts as an auxiliary visual reference, providing self-feedback to the LVLM to verify and correct its initial response, often through complementary or contrastive decoding techniques.

    \item \textbf{Woodpecker}~\cite{yin2023woodpecker}: Woodpecker is a post-hoc correction framework designed to mitigate hallucinations in the outputs of Multimodal Large Language Models (MLLMs). It operates by first prompting the MLLM itself to identify potential hallucinations (across several predefined types like object existence, attributes, etc.) in its initial response. If hallucinations are detected, Woodpecker then instructs the MLLM to revise and correct these identified errors.

    \item \textbf{HALC}~\cite{chen2024halc}: HALC (e.g., "Mitigating Object Hallucinations in Large Vision-Language Models via Cause Analysis and Post-hoc Correction") is a post-hoc method that first analyzes the potential causes of object hallucinations to identify objects in the response that are likely to be hallucinated. Subsequently, it instructs the LVLM to verify the existence of these specific, suspect objects within the image and make corrections if they are indeed confirmed as hallucinations.

    \item \textbf{OPERA}~\cite{huang2024opera}: OPERA aims to alleviate visual relation hallucinations in LVLMs during decoding. It consists of two main components: an Over-trust Penalty (OP) term designed to penalize the model's over-confidence on unreliable visual relations during token generation, and a Retrospection-Allocation (RA) mechanism that encourages the model to retrospect previously generated tokens and re-allocate attention to relevant visual regions for verification and potential correction.
\end{itemize}
We report the performance of these baselines based on our re-implementation using their released code bases where available.

\section{Implementation Details}
\label{sec:appendix_implementation_details}
In all experiments using MRFD, we set the number of regions $K=3$ to obtain cropped images and the temperature $\gamma=0.02$ for JSD-based weighting (Eq.~\ref{eq:weight}). For the decoding process, we employ multinomial sampling in both stages. Specifically, during the first step to generate the initial analyses $r_k$, we use a sampling temperature of 0.7. In the second step to generate the final output sequence $y$, we use a lower sampling temperature of 0.1 after fusing the logits.

To obtain the aggregated spatial attention map $\hat{A}$ (Eq.~\ref{eq:attention_reshape}), we first average the attention weights across all attention heads within the relevant layer(s). For \textbf{LLaVA-1.5}, we utilize the attention map from the final cross-attention layer. For \textbf{InstructBLIP}, which uses Q-Former, we identify the query token with the maximum aggregated attention score and use its corresponding attention map. The spatial dimension $d'$ of the map $\hat{A}$ corresponds to the grid size of the visual patches ($m=d' \times d'$), which is $24 \times 24$ for LLaVA-1.5 (ViT-L-336px) and $16 \times 16$ for the ViT-g/14 used by InstructBLIP.

To efficiently identify the top-K salient regions $\{R_1, ..., R_K\}$ based on $\hat{A}$, we employ an integral image approach \cite{viola2001rapid} to quickly calculate the sum of attention scores within any rectangular bounding box. We search for the K non-overlapping or minimally overlapping rectangular regions that maximize these summed attention scores. To ensure diversity in the selected regions, we enforce a maximum Intersection over Union (IoU) of 40\% between any pair of selected bounding boxes $R_i$ and $R_j$ ($i \neq j$).

\section{Datasets and Benchmarks}
\label{sec:data}

We evaluate our MRFD framework on a diverse set of benchmarks targeting both hallucination detection and general vision-language capabilities.
\begin{itemize}
\item\textbf{POPE \cite{li2023evaluating_pope}:}
POPE (Polling-based Object Probing Evaluation) is a widely used benchmark for assessing object existence hallucination in LVLMs. It presents models with Yes/No questions concerning the presence of specific objects (e.g., "Is there a \{object\} in the image?"). The benchmark data is structured into three main subsets derived from MSCOCO~\cite{lin2014microsoft_coco}, A-OKVQA~\cite{schwenk2022okvqa}, and GQA~\cite{hudson2019gqa}. Each of these subsets is further divided based on three negative sampling strategies for non-existent objects: \textit{random}, \textit{popular}, and \textit{adversarial}, which vary in difficulty. For evaluation, we report standard metrics including Accuracy, Precision, Recall, and F1 score. 

\item\textbf{MME \cite{fu2023mme}:}
MME serves as a comprehensive benchmark for evaluating overall LVLM perception and cognition. Our evaluation specifically utilizes the \textbf{MME-Hallucination} subset, which is designed to assess a range of common hallucination types. These are categorized into object-level assessments like \textit{existence} and \textit{count}, and attribute-level assessments such as object \textit{position} and \textit{color}. Questions in this subset are typically Yes/No queries. We report scores based on the official benchmark protocol, which often involves combined accuracy measures reflecting both question-level and image-level correctness. 

\item\textbf{CHAIR \cite{rohrbach2018object_chair}:}
The CHAIR (Caption Hallucination Assessment with Image Relevance) benchmark quantifies object hallucinations within the context of open-ended image captioning. LVLMs are prompted to generate descriptive captions for images, for which we, following prior work~\cite{lee2024volcano}, use a random selection of 500 images from the MSCOCO~\cite{lin2014microsoft_coco} validation set. The generated captions are then compared against ground-truth objects within the image to calculate the CHAIRi (instance-level) and CHAIRs (category-level) scores, where lower scores indicate fewer hallucinations.

\begin{equation}
    \begin{split}
        &\text{CHAIR}_{S}=\frac{\text{captions w/ hallucinated objects}}{\text{all captions}}, \\&\text{CHAIR}_{I}=\frac{\text{hallucinated objects}}{\text{all mentioned objects}}.
    \end{split}
  \end{equation}

\item\textbf{MMBench \cite{liu2023mmbench}:}
MMBench evaluates a broad spectrum of multimodal capabilities through carefully curated multiple-choice questions that span various cognitive dimensions and skills. For this benchmark, we adhere to the official evaluation protocol and report the overall accuracy score.
\end{itemize}

\section{Detailed Results of experiments}
\label{sec:detailed_experiments}

\subsection{Detailed Results of CHAIR}
\label{detailed_chair}
Detailed results of CHAIR are shown in Table~\ref{tab:chair_128} and Table~\ref{tab:chair_256}, reporting CHAIRs (Cs) and CHAIRi (Ci) scores (lower is better) for LLaVA-1.5 and InstructBLIP.  MRFD consistently achieves state-of-the-art performance on both LVLMs.
\begin{table}[H]
    \centering

    \begin{tabular}{lccccc}
    \toprule
    \multirow{2}{*}{{Method}} & \multicolumn{2}{c}{{LLaVA-1.5}} & \multicolumn{2}{c}{{InstructBLIP}} \\
    \cmidrule(lr){2-3} \cmidrule(lr){4-5}
    & Cs $\downarrow$ & Ci $\downarrow$ & Cs $\downarrow$ & Ci $\downarrow$ \\
    \midrule
    Regular    & 55.0 & 16.3 & 57.0 & 17.6 \\
    VCD        & 54.4 & 16.6 & 60.4 & 17.8 \\
    M3ID       & 56.6 & 15.7 & 62.2 & 18.1 \\
    RITUAL     & 49.6 & 14.8 & \underline{48.4} & 14.5 \\
    Woodpecker & 57.6 & 16.7 & 60.8 & 17.6 \\
    HALC       & 51.0 & 14.8 & 53.8 & 15.7 \\
    DeGF       & \underline{48.8} & \underline{14.6}  & 49.2 & \underline{14.4} \\
    \midrule
    \textbf{Ours (MRFD)} & \textbf{37.1} & \textbf{9.2}  & \textbf{38.2} & \textbf{10.9} \\
    \bottomrule
    \end{tabular}
    \caption{Results on CHAIR benchmark for caption generation. We limited the maximum number of new tokens to 128. Lower ($\downarrow$) CHAIRs (Cs) and CHAIRi (Ci) indicate less hallucination. Best results are bolded, second-best are underlined.}
    \label{tab:chair_128}
\end{table}

\begin{table}[t]
    \centering

    \begin{tabular}{lccccc}
    \toprule
    \multirow{2}{*}{\textbf{Method}} & \multicolumn{2}{c}{\textbf{LLaVA-1.5}} & \multicolumn{2}{c}{\textbf{InstructBLIP}} \\
    \cmidrule(lr){2-3} \cmidrule(lr){4-5}
    & Cs $\downarrow$ & Ci $\downarrow$ & Cs $\downarrow$ & Ci $\downarrow$ \\
    \midrule
    Regular    & 58.0          & 17.7           & 61.0          & 18.2 \\
    VCD        & 58.2          & 16.7           & 63.0          & 18.6 \\
    M3ID       & 56.8 & 16.1 & 65.8          & 19.9 \\
    RITUAL     & 51.0          & 15.1           & 50.4         & 15.3  \\
    DeGF       & \underline{49.8}          & \underline{14.7}           & \underline{49.8}          & \underline{15.1}  \\
    \midrule
    \textbf{Ours (MRFD)} & \textbf{39.0} & \textbf{11.0}  & \textbf{38.6} & \textbf{11.3} \\
    \bottomrule
    \end{tabular}
    \caption{Results on CHAIR benchmark for caption generation. We limited the maximum number of new tokens to 256. Lower ($\downarrow$) CHAIRs (Cs) and CHAIRi (Ci) indicate less hallucination. Best results are bolded, second-best are underlined.}
    \label{tab:chair_256}
\end{table}

\subsection{Detailed Results of MME}
\label{detailed_mme}
\begin{figure}[H]
    \centering
    \includegraphics[width=1\columnwidth]{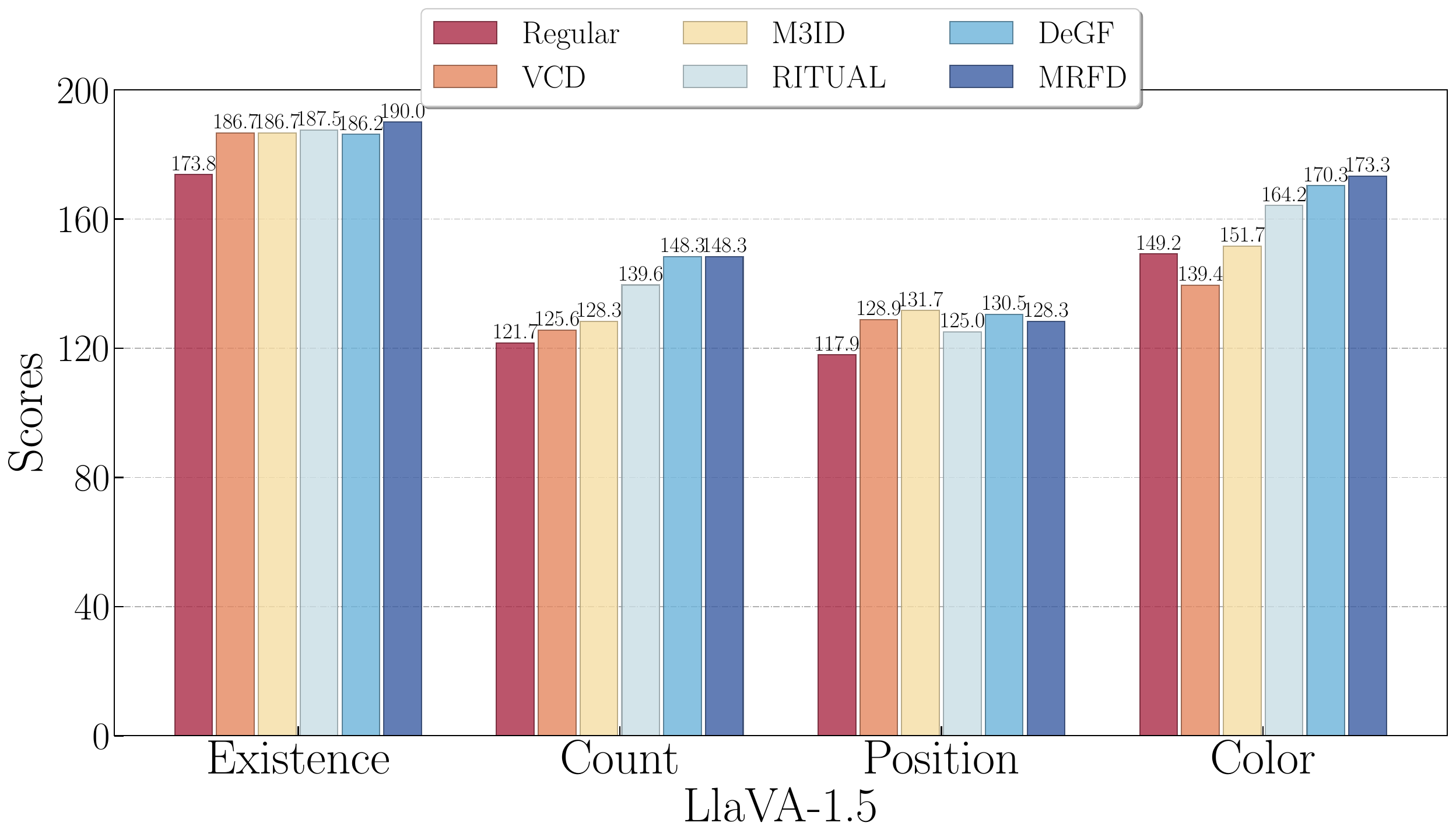}
    \includegraphics[width=1\columnwidth]{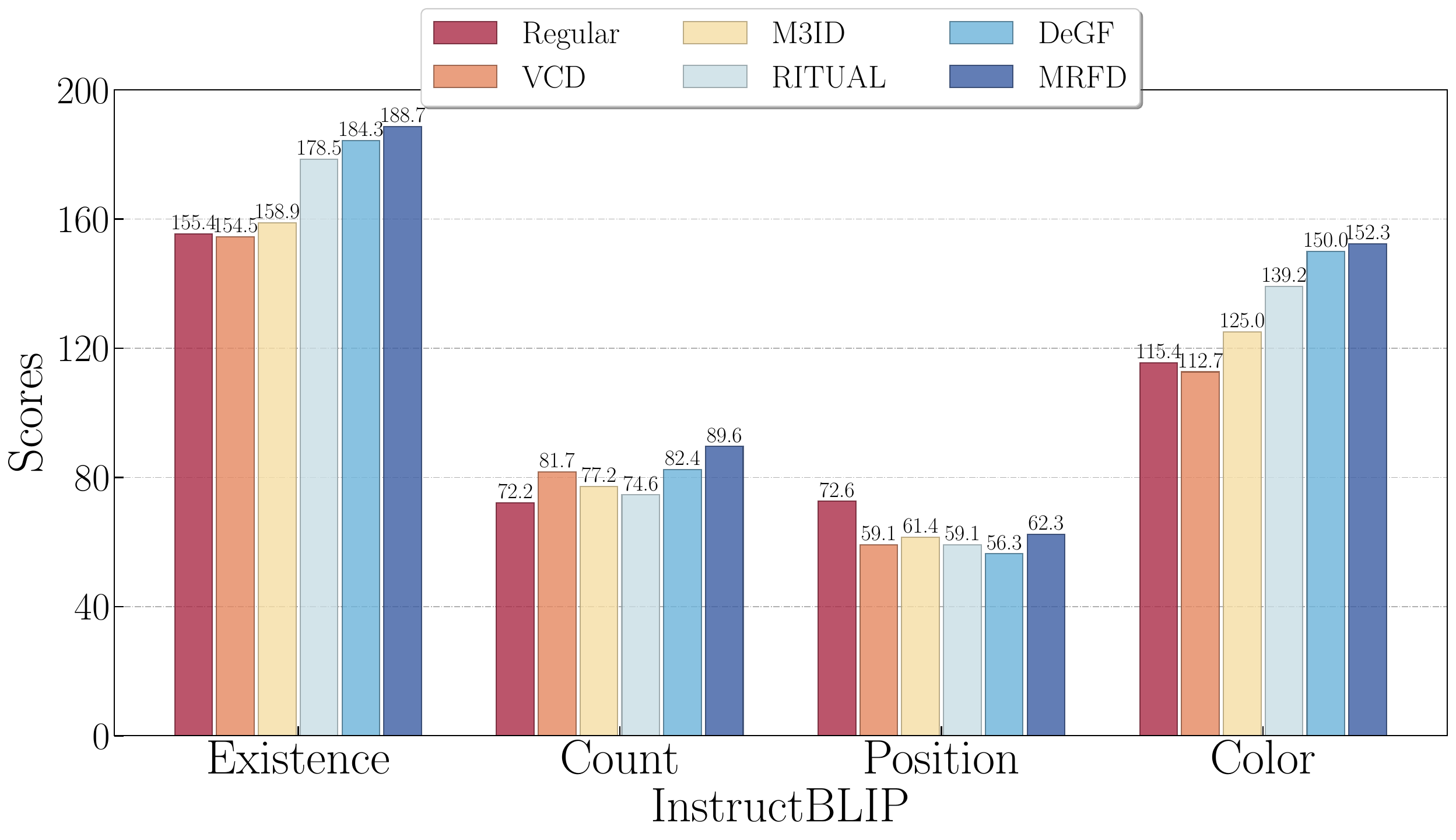}
    \caption{Experimental results of MME with different decoding strategies.}
    \label{fig:mme_detailed}
\end{figure}

In table~\ref{tab:mme_detailed_benchmark_results} and Figure~\ref{fig:mme_detailed}, we provide detailed results on the MME-Hallucination benchmark~\cite{fu2023mme} for both LLaVA-1.5 and InstructBLIP. The table includes scores for object-level (existence, count) and attribute-level (position, color) tasks, and averages the score across three random seeds. The best results are bolded, and the second-best are underlined.

\subsection{Detailed Results of POPE}
\label{detailed_pope}
In table~\ref{tab:detailed_pope}, we provide detailed results on the POPE benchmark~\cite{li2023evaluating_pope} for both LLaVA-1.5 and InstructBLIP across three different settings (Random, Popular, Adversarial). The table includes accuracy (Acc.), precision (Prec.), recall, and F1 scores for each method. The best results are bolded, and the second-best are underlined.

\begin{table*}[ht]
    \centering
    \small
    \resizebox{\textwidth}{!}{%
    \begin{tabular}{lllcccccccc}
    \toprule
    \multirow{2}{*}{\textbf{Dataset}} & \multirow{2}{*}{\textbf{Setting}} & \multirow{2}{*}{\textbf{Method}} & \multicolumn{4}{c}{\textbf{LLaVA-1.5}} & \multicolumn{4}{c}{\textbf{InstructBLIP}} \\
    \cmidrule(lr){4-7} \cmidrule(lr){8-11}
    & & & Acc. $\uparrow$ & Prec. $\uparrow$ & Recall $\uparrow$ & F1 $\uparrow$ & Acc. $\uparrow$ & Prec. $\uparrow$ & Recall $\uparrow$ & F1 $\uparrow$ \\
    \midrule
    \multirow{18}{*}{MS-COCO}
    & \multirow{6}{*}{Random}      & Regular & 83.13 & 81.94 & 85.00 & 83.44 & 83.07 & 83.02 & 83.26 & 83.08 \\
    &                              & VCD     & 87.00 & 86.13 & \underline{88.18} & 87.15 & 86.23 & 88.14 & 83.73 & 85.88 \\
    &                              & M3ID    & 87.50 & 87.38 & 87.67 & 87.52 & 86.67 & 88.09 & \underline{84.79} & 86.41 \\
    &                              & RITUAL  & 88.87 & 89.23 & \textbf{88.39} & \underline{88.81} & \textbf{88.83} & 90.48 & \textbf{86.80} & \textbf{88.60} \\
    &                              & DeGF    & \underline{89.03} & \underline{91.20} & 86.41 & 88.74 & \textbf{88.83} & \underline{93.73} & 82.42 & 87.71 \\
    \rowcolor{green!26} & & \cellcolor{green!26}{Ours} & \textbf{89.50} & \textbf{92.55} & 85.94 & \textbf{89.12} & \underline{88.52} & \textbf{93.83} & 82.61 & \underline{87.86} \\
    \cmidrule(lr){2-11}
    & \multirow{6}{*}{Popular}     & Regular & 81.17 & 78.28 & 86.26 & 82.08 & 77.00 & 73.82 & 83.68 & 78.44 \\
    &                              & VCD     & 83.10 & 79.96 & \underline{88.34} & 83.94 & 80.07 & 77.67 & 84.39 & 80.89 \\
    &                              & M3ID    & 84.30 & 81.58 & \textbf{88.63} & 84.95 & 80.97 & 77.93 & \underline{86.19} & 81.85 \\
    &                              & RITUAL  & 85.83 & 84.17 & 88.26 & 86.17 & 81.97 & 78.90 & \textbf{87.26} & \underline{82.87} \\
    &                              & DeGF    & \underline{86.63} & \textbf{87.75} & 84.86 & \underline{86.28} & \underline{82.73} & \underline{84.02} & 80.27 & 82.10 \\
    \rowcolor{green!26} & & \cellcolor{green!26}{Ours} & \textbf{87.24} & \underline{86.56} & 88.22 & \textbf{87.38} & \textbf{83.69} & \textbf{85.22} & 81.58 & \textbf{83.36} \\
    \cmidrule(lr){2-11}
    & \multirow{6}{*}{Adversarial} & Regular & 77.43 & 73.31 & 86.26 & 79.26 & 74.60 & 71.26 & 82.46 & 76.45 \\
    &                              & VCD     & 77.17 & 72.18 & \textbf{88.40} & 79.47 & 77.20 & 74.29 & 83.19 & 78.49 \\
    &                              & M3ID    & 78.23 & 73.51 & \underline{88.28} & 80.22 & 77.47 & 73.68 & \underline{85.48} & 79.14 \\
    &                              & RITUAL  & 78.80 & 74.43 & 87.74 & 80.54 & 78.73 & 74.57 & \textbf{87.21} & \underline{80.39} \\
    &                              & DeGF    & \underline{81.63} & \underline{80.59} & 83.33 & \underline{81.94} & \underline{80.30} & \underline{80.90} & 79.33 & 80.11 \\
    \rowcolor{green!26} & & \cellcolor{green!26}{Ours} & \textbf{82.75} & \textbf{85.22} & 79.25 & \textbf{82.12} & \textbf{82.49} & \textbf{83.14} & 81.51 & \textbf{82.32} \\
    \midrule
    \multirow{18}{*}{A-OKVQA}
    & \multirow{6}{*}{Random}      & Regular & 81.90 & 76.63 & 91.78 & 83.53 & 80.63 & 76.82 & 87.75 & 81.92 \\
    &                              & VCD     & 83.83 & 78.05 & \underline{94.13} & 85.34 & 84.20 & 80.90 & 89.54 & 85.00 \\
    &                              & M3ID    & 84.67 & 79.25 & 93.94 & 85.97 & 85.43 & 81.77 & \underline{91.20} & 86.23 \\
    &                              & RITUAL  & 85.17 & 79.79 & \textbf{94.21} & 86.40 & 87.13 & 83.92 & \textbf{91.87} & \underline{87.71} \\
    &                              & DeGF    & \underline{86.93} & \underline{84.28} & 90.80 & \textbf{87.42} & \underline{87.40} & \textbf{87.67} & 86.86 & 87.26 \\
    \rowcolor{green!26} & & \cellcolor{green!26}{Ours} & \textbf{87.13} & \textbf{87.56} & 86.76 & \underline{87.16} & \textbf{88.33} & \underline{86.81} & 90.40 & \textbf{88.57} \\
    \cmidrule(lr){2-11}
    & \multirow{6}{*}{Popular}     & Regular & 75.07 & 68.58 & 92.53 & 78.77 & 75.17 & 70.15 & 87.60 & 77.91 \\
    &                              & VCD     & 76.63 & 69.59 & \textbf{94.59} & 80.19 & 78.63 & 73.53 & 89.46 & 80.72 \\
    &                              & M3ID    & 77.80 & 70.98 & 94.07 & 80.91 & 78.80 & 73.38 & 90.39 & 81.00 \\
    &                              & RITUAL  & 78.83 & 71.99 & \underline{94.37} & 81.68 & 78.73 & 72.83 & \textbf{91.68} & 81.17 \\
    &                              & DeGF    & \underline{80.90} & \underline{75.68} & 91.05 & \textbf{82.66} & \underline{81.47} & \underline{78.61} & 86.47 & \underline{82.35} \\
    \rowcolor{green!26} & & \cellcolor{green!26}{Ours} & \textbf{80.99} & \textbf{76.68} & 89.28 & \underline{82.51} & \textbf{83.19} & \textbf{78.74} & \underline{91.00} & \textbf{84.43} \\
    \cmidrule(lr){2-11}
    & \multirow{6}{*}{Adversarial} & Regular & 67.23 & 61.56 & 91.81 & 73.70 & 69.87 & 64.54 & 88.20 & 74.54 \\
    &                              & VCD     & 67.40 & 61.39 & 93.79 & 74.21 & 71.00 & 65.41 & 89.13 & 75.45 \\
    &                              & M3ID    & 68.60 & 62.22 & \textbf{94.74} & 75.11 & 70.10 & 64.28 & 90.47 & 75.16 \\
    &                              & RITUAL  & 68.57 & 62.26 & \underline{94.27} & 74.99 & 70.27 & 64.15 & \textbf{91.89} & 75.55 \\
    &                              & DeGF    & \underline{72.70} & \underline{66.70} & 90.68 & \underline{76.86} & \underline{73.93} & \underline{69.36} & 85.70 & \underline{76.67} \\
    \rowcolor{green!26} & & \cellcolor{green!26}{Ours} & \textbf{75.23} & \textbf{71.03} & 85.39 & \textbf{77.65} & \textbf{75.62} & \textbf{69.40} & \underline{91.78} & \textbf{79.04} \\
    \midrule
    \multirow{18}{*}{GQA}
    & \multirow{6}{*}{Random}      & Regular & 82.23 & 76.32 & 93.47 & 84.03 & 79.67 & 76.05 & 86.62 & 80.99 \\
    &                              & VCD     & 83.23 & 76.73 & \underline{95.38} & 85.05 & 82.83 & 80.16 & 87.26 & 83.56 \\
    &                              & M3ID    & 84.20 & 78.00 & 95.26 & 85.77 & 83.07 & 80.06 & 88.06 & 83.87 \\
    &                              & RITUAL  & 86.10 & 80.30 & \textbf{95.66} & 87.31 & 84.87 & 82.52 & \underline{88.47} & \underline{85.39} \\
    &                              & DeGF    & \underline{87.40} & \underline{83.51} & 93.20 & \underline{88.09} & \underline{85.40} & \textbf{85.64} & 84.61 & 85.12 \\
    \rowcolor{green!26} & & \cellcolor{green!26}{Ours} & \textbf{87.81} & \textbf{86.62} & 90.32 & \textbf{88.41} & \textbf{87.24} & \underline{85.57} & \textbf{89.69} & \textbf{87.58} \\
    \cmidrule(lr){2-11}
    & \multirow{6}{*}{Popular}     & Regular & 73.47 & 66.83 & 93.20 & 77.84 & 73.33 & 68.72 & 85.67 & 76.26 \\
    &                              & VCD     & 72.37 & 65.27 & \underline{95.60} & 77.58 & 76.13 & 71.10 & 88.07 & \underline{78.68} \\
    &                              & M3ID    & 73.87 & 66.70 & 95.35 & 78.49 & 75.17 & 69.94 & 88.26 & 78.04 \\
    &                              & RITUAL  & 74.80 & 67.50 & \textbf{95.66} & 79.15 & 74.50 & 69.17 & \underline{88.39} & 77.61 \\
    &                              & DeGF    & \textbf{78.10} & \underline{71.56} & 93.25 & \textbf{80.98} & \underline{76.90} & \underline{73.89} & 83.20 & 78.27 \\
    \rowcolor{green!26} & & \cellcolor{green!26}{Ours} & \underline{77.72} & \textbf{73.17} & 88.24 & \underline{80.00} & \textbf{79.62} & \textbf{74.61} & \textbf{90.15} & \textbf{81.65} \\
    \cmidrule(lr){2-11}
    & \multirow{6}{*}{Adversarial} & Regular & 68.60 & 62.43 & 93.41 & 74.84 & 68.60 & 63.94 & 85.31 & 73.10 \\
    &                              & VCD     & 68.83 & 62.26 & \underline{95.67} & 75.43 & 71.00 & 65.75 & 87.66 & 75.14 \\
    &                              & M3ID    & 68.67 & 62.16 & 95.42 & 75.28 & 71.17 & 65.79 & \underline{88.19} & \underline{75.36} \\
    &                              & RITUAL  & 68.23 & 61.75 & \textbf{95.81} & 75.10 & 70.17 & 64.76 & \textbf{88.48} & 74.78 \\
    &                              & DeGF    & \underline{74.07} & \underline{67.42} & 93.14 & \textbf{78.22} & \underline{73.63} & \underline{70.08} & 80.92 & 75.11 \\
    \rowcolor{green!26} & & \cellcolor{green!26}{Ours} & \textbf{76.00} & \textbf{72.22} & 84.50 & \underline{77.88} & \textbf{75.06} & \textbf{70.25} & 87.18 & \textbf{77.80} \\
    \bottomrule
    \end{tabular}
    }
    \caption{Detailed results on POPE (Li et al., 2023d) benchmark. Higher ($\uparrow$) accuracy, precision, recall, and F1 indicate better performance. The best results are bolded, and the second-best are underlined.}
    \label{tab:detailed_pope}
\end{table*}

\begin{table*}[b]
\centering

\begin{tabular}{@{}llccccc@{}}
\toprule
\multirow{2}{*}{\textbf{Model}} & \multirow{2}{*}{\textbf{Method}} & \multicolumn{2}{c}{\textbf{Object-level}} & \multicolumn{2}{c}{\textbf{Attribute-level}} & \multirow{2}{*}{\textbf{Total Score $\uparrow$}} \\
\cmidrule(lr){3-4} \cmidrule(lr){5-6}
& & \textbf{Existence $\uparrow$} & \textbf{Count $\uparrow$} & \textbf{Position $\uparrow$} & \textbf{Color $\uparrow$} & \\
\midrule
\multirow{7}{*}{LLaVA-1.5} & Regular & 173.75 & 121.67 & 117.92 & 149.17 & 562.50 \\
& VCD & 186.67 & 125.56 & 128.89 & 139.45 & 580.56 \\
& M3ID & 186.67 & 128.33 & \underline{131.67} & 151.67 & 598.11 \\
& RITUAL & 187.50 & \underline{139.58} & 125.00 & \underline{164.17} & \underline{616.25} \\
& DeGF & 186.22 & 148.33 & 130.50 & 170.33 & 635.38 \\ 
\rowcolor{green!26}  \cellcolor{white} & Ours & \textbf{190.00} & \textbf{148.33} & \textbf{128.33} & \textbf{173.33} & \textbf{640.00} \\ 
\midrule
\multirow{6}{*}{InstructBLIP} & Regular & 155.42 & 72.17 & \textbf{72.58} & 115.43 & 415.60 \\
& VCD & 154.49 & 81.67 & 59.11 & 112.67 & 407.94 \\
& M3ID & 158.89 & 77.22 & 61.44 & 125.00 & 422.55 \\
& RITUAL & 178.50 & 74.58 & 59.08 & 139.17 & 451.33 \\
& DeGF & \underline{184.32} & \underline{82.44} & 56.33 & \underline{150.00} & \underline{473.09} \\
\rowcolor{green!26}  \cellcolor{white} & Ours & \textbf{188.67} & \textbf{89.58} & \underline{62.33} & \textbf{152.33} & \textbf{492.91} \\
\bottomrule
\end{tabular}
\caption{Detailed performance on the MME-Hallucination benchmark. Scores are reported as mean. Higher scores $\uparrow$ indicate better performance. For each model group LLaVA-1.5, InstructBLIP, results for "Ours" are \textbf{bolded} if they are the best in that column. Other best results in a column are also \textbf{bolded}. Underlined values \underline{score} typically represent the second-best performing method or a notable baseline. The "Ours" rows are highlighted.}
\label{tab:mme_detailed_benchmark_results}
\end{table*}

\section{Efficiency Comparison}
\label{detailed_efficiency}
We report MRFD's efficiency on the CHAIR benchmark with two backbones—InstructBLIP-7B and LLaVA-1.5-7B—under the same decoding setup (128 max tokens, RTX 3090 GPU). 
With multi-region ($K{=}3$) analysis and fusion, MRFD increases latency by about ${\sim}2.84\times$ (from 3.59\,s to 10.21\,s) and peak GPU memory by ${\sim}1.11\times$ (from 15{,}238\,MB to 16{,}932\,MB) relative to Regular decoding on InstructBLIP-7B (Table~\ref{tab:efficiency-blip}). 
On LLaVA-1.5-7B, MRFD shows a ${\sim}3.30\times$ latency increase (3.44\,s $\rightarrow$ 11.34\,s) and a ${\sim}1.11\times$ memory increase (15{,}778\,MB $\rightarrow$ 17{,}458\,MB) over Regular (Table~\ref{tab:efficiency-llava}).
Despite this overhead—and compared to lighter baselines such as VCD or the post-hoc Woodpecker—MRFD remains markedly more efficient than iterative/feedback-heavy methods (DeGF, OPERA, HALC), while achieving the best CHAIRs (lower is better): 38.2 on InstructBLIP-7B and 37.1 on LLaVA-1.5-7B, representing ${\sim}33\%$ and ${\sim}32.5\%$ reductions from their respective Regular baselines. 
Overall, MRFD offers a compelling balance between computational cost and hallucination reduction.

\begin{table}[H]
    \centering
    \small
    \begin{tabular}{lccc}
    \toprule
    \textbf{Method} & \textbf{Avg. Lat. $\downarrow$} & \textbf{GPU Mem. $\downarrow$} & \textbf{CHAIRs $\downarrow$} \\
    \midrule
    Regular    & 3.59 s  & 15{,}238 MB  & 57.0 \\
    VCD        & 6.82 s  & 16{,}078 MB  & 60.4 \\
    OPERA      & 25.31 s & 23{,}575 MB  & 62.2 \\
    Woodpecker & 11.72 s & 22{,}798 MB  & 60.8 \\
    HALC       & 21.84 s & 22{,}689 MB  & 53.8 \\
    DeGF       & 14.69 s & 18{,}726 MB  & 49.2 \\
    \midrule
    \textbf{Ours (MRFD)} & \textbf{10.21 s} & \textbf{16{,}932 MB} & \textbf{38.2} \\
    \bottomrule
    \end{tabular}
    \caption{Efficiency comparison on \textbf{InstructBLIP-7B}. Lower is better for all metrics.}
    \label{tab:efficiency-blip}
\end{table}

\begin{table}[H]
    \centering
    \small
    \begin{tabular}{lccc}
    \toprule
    \textbf{Method} & \textbf{Avg. Lat. $\downarrow$} & \textbf{GPU Mem. $\downarrow$} & \textbf{CHAIRs $\downarrow$} \\
    \midrule
    Regular    & 3.44 s  & 15{,}778 MB  & 55.0 \\
    VCD        & 6.91 s  & 16{,}634 MB  & 54.4 \\
    OPERA      & 24.70 s & 22{,}706 MB  & 52.6 \\
    Woodpecker & 10.68 s & 22{,}199 MB  & 57.6 \\
    HALC       & 22.61 s & 23{,}084 MB  & 51.0 \\
    DeGF       & 13.89 s & 19{,}119 MB  & 48.8 \\
    \midrule
    \textbf{Ours (MRFD)} & \textbf{11.34 s} & \textbf{17{,}458 MB} & \textbf{37.1} \\
    \bottomrule
    \end{tabular}
    \caption{Efficiency comparison on \textbf{LLaVA-1.5-7B}. Lower is better for all metrics.}
    \label{tab:efficiency-llava}
\end{table}

\section{Detialed Ablation Study}
\label{detailed_ablation}
\subsection{Components Ablation}
To assess the contributions of MRFD's core components—attention-guided Region Selection (RS), JSD-based Consistency Weighting (CW), and Fusion Prompts (FP)—we conduct ablation studies. Results on POPE-COCO (average all settings, LLaVA-1.5) are in Table~\ref{tab:detailed_ablation_components}. The evaluated model variants are:

\begin{itemize}
    \item[(a)] \textbf{MRFD (Full)} : Our complete method, integrating all three components.
    \item[(b)] \textbf{w/o CW} : Employs RS and FP, but uses uniform weights for fusing predictions from multiple regions, bypassing JSD-based consistency weighting.
    \item[(c)] \textbf{w/o FP} : Utilizes RS and CW, but omits the enriched Fusion Prompts, using only the original question for each regional decoding pass.
    \item[(d)] \textbf{w/o RS (Global Image + FP)}: Bypasses attention-guided region selection, operating solely on the global image. The Fusion Prompt is constructed based on the global image's initial response. In this single-view context, the multi-region Consistency Weighting (CW) mechanism as defined is not applicable or becomes trivial (effectively a weight of 1 for the single view).
    \item[(e)] \textbf{RS only} : Leverages attention-guided regions, but with uniform fusion weights and only the original question as prompt, isolating the benefit of the multi-region perspective itself.
    \item[(f)] \textbf{FP only (Global Image)} : Applies the Fusion Prompt (derived from the global image's initial response) directly to the global image decoding, without multi-region selection or any form of consistency weighting.
    \item[(g)] \textbf{Global Image (Token-Level Voting)} : Operates on the global image with the original prompt. The CW component is adapted to perform token-level voting by aggregating implicitly diversified predictions from the single global view, testing self-consistency benefits at the token level.
    \item[(h)] \textbf{Regular} : Standard greedy decoding using only the global image and the original question.
\end{itemize}

\begin{table}[H]
    \centering
        \begin{tabular}{lcccccc}
        \toprule
        \multirow{2}{*}{} & \multicolumn{3}{c}{Components} & \multirow{2}{*}{Acc. $\uparrow$} & \multirow{2}{*}{Prec. $\uparrow$} & \multirow{2}{*}{F1 $\uparrow$} \\
        \cmidrule(lr){2-4}
        & CW & FP & RS & & & \\
        \midrule
        (a) & \checkmark & \checkmark & \checkmark & \textbf{86.50} & \textbf{88.11} & \textbf{86.21} \\
        \midrule
        (b)  &  & \checkmark & \checkmark & 83.76 & 84.27 & 83.74 \\ 
        (c)  & \checkmark &  & \checkmark & 82.87 & 83.95 & 82.70 \\ 
        (d)  & \checkmark & \checkmark &  & 82.77 & 83.94 & 82.58 \\ 
        (e) &  &  & \checkmark & 82.58 & 82.01 & 82.26 \\
        (f) &  & \checkmark &  & 81.96 & 81.36 & 82.14 \\
        (g) & \checkmark &  &  & 81.52 & 81.05 & 81.68 \\
        (h)  &  &  &  & 80.57 & 78.84 & 81.59 \\
        \bottomrule
        \end{tabular}
        \caption{Ablation study with different model variants on POPE-COCO under the average of three settings. CW: JSD-based Consistency Weighting, FP: Fusion Prompt, RS: Region Selection.}
        \label{tab:detailed_ablation_components}
        \end{table}

\subsection{Decoding Strategies}
To evaluate the impact of different decoding strategies on MRFD, we conduct an ablation study using the POPE-COCO benchmark under the adversarial setting with LLaVA-1.5. The results are summarized in Table~\ref{tab:ablation_decoding_strategies}. We compare three decoding strategies: low temperature sampling, high temperature sampling, and high temperature sampling with top-p filtering. The best results are bolded.

The low temperature strategy is the default setting in our experiments, which $t$ is 0.1. The high temperature strategy is set to $t$=0.5, which increases the randomness of the sampling process. The high temperature + top-p strategy combines high temperature sampling with top-p filtering, where we set $p$=0.7 to retain the top 70\% of the probability mass.
            \begin{table}[H]
            \centering
            \begin{tabular}{@{}l|ccc@{}}
            \toprule
            Decoding  & Acc. $\uparrow$& Prec. $\uparrow$& F1 $\uparrow$\\
            \midrule
            Low temp.  & 82.75 & 85.22 & 82.12 \\
            \midrule
            High Temp.  & 83.04 & 82.29 & 83.23 \\
            \midrule
            High Temp.+Top P  & 82.50 & 81.25 & 82.67 \\
            \bottomrule
            \end{tabular}
            \caption{Ablation study on different decoding strategies.}
            \label{tab:ablation_decoding_strategies}
            \end{table}

\section{Qualitative Examples}
\label{sec:appendix_qualitative_chair}

\begin{figure*}[t]
\centering
\includegraphics[width=1.0\linewidth]{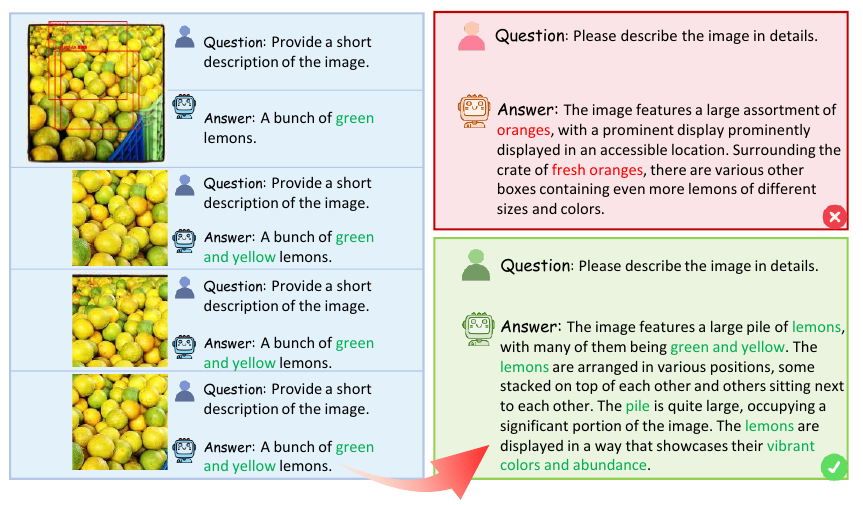} 
\includegraphics[width=1.0\linewidth]{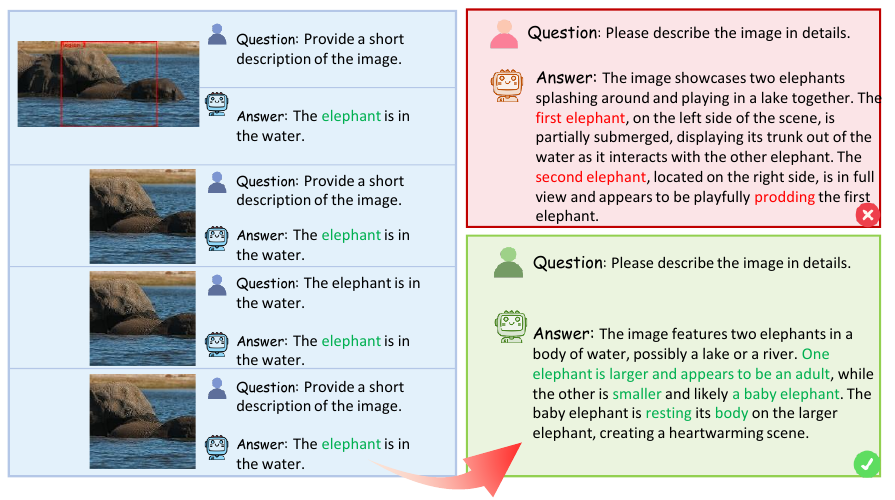} 
\caption{Qualitative comparison of captions generated by Regular decoding (Red) and our MRFD (Green) on an image from the CHAIR evaluation set. }
\label{fig:qualitative_results1} 
\end{figure*}

\begin{figure*}[t]
\centering
\includegraphics[width=1.0\linewidth]{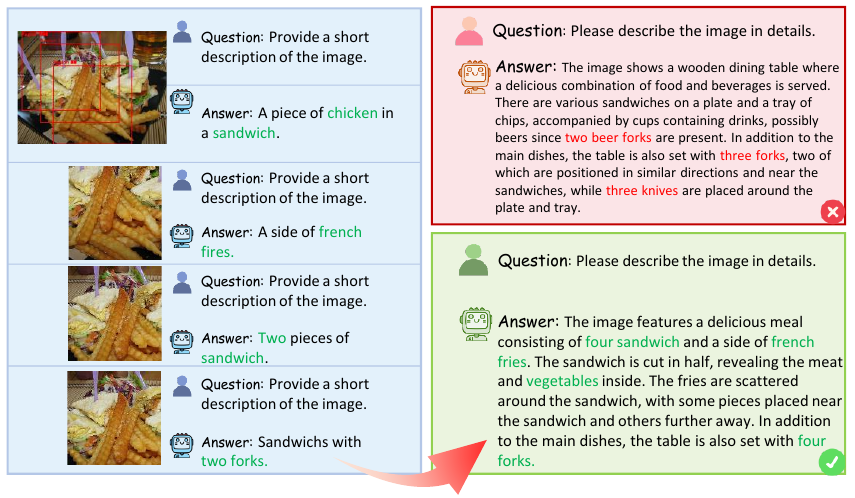} 
\includegraphics[width=1.0\linewidth]{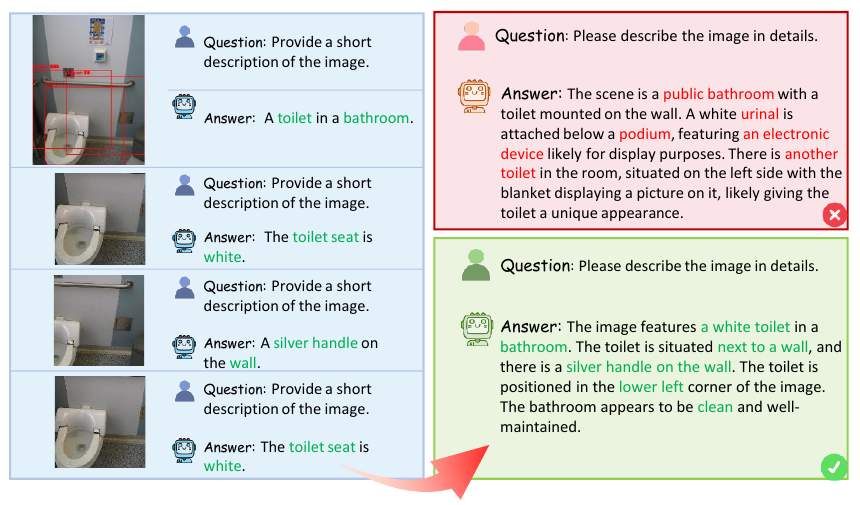} 
\caption{Qualitative comparison of captions generated by Regular decoding (Red) and our MRFD (Green) on an image from the CHAIR evaluation set.}
\label{fig:qualitative_results2} 
\end{figure*}

To provide a more intuitive understanding of MRFD's impact on mitigating hallucinations and improving caption factuality, this section presents qualitative examples from our experiments on the CHAIR benchmark using LlaVA-1.5. We compare captions generated by our Multi-Region Fusion Decoding (MRFD) method against those from the standard Regular baseline. These examples are selected to highlight differences in how the methods perceive and describe objects, attributes, and their relationships within the visual scene. Figure~\ref{fig:qualitative_results1} and Figure~\ref{fig:qualitative_results2} illustrate one such comparison. 

\end{document}